\documentclass[11pt]{article}

\usepackage{acl}

\usepackage{times}
\usepackage{latexsym}
\usepackage[T1]{fontenc}
\usepackage[utf8]{inputenc}
\usepackage{microtype}
\usepackage{inconsolata}
\usepackage{graphicx}
\usepackage{amsmath}
\usepackage{amssymb}
\usepackage{booktabs}
\usepackage{multirow}
\usepackage{xcolor}
\usepackage{subcaption}
\usepackage{tikz}
\usetikzlibrary{patterns,positioning,arrows.meta,calc,shapes.geometric,fit,decorations.pathreplacing}
\usepackage{pgfplots}

\usepackage{hyperref}
\hypersetup{
	pdftitle={Sparse Autoencoders Map Brain–LLM Alignment onto Cortical Semantic Topography},
	pdfauthor={Dongxin Guo, Jikun Wu, Siu Ming Yiu},
	pdfsubject={Computational Neurolinguistics; CoNLL 2026},
	pdfkeywords={sparse autoencoders, mechanistic interpretability, brain–LLM alignment, cortical semantic topography, fMRI, neural encoding, sparse coding, GPT-2, Llama, naturalistic language}
}
\pgfplotsset{compat=1.18}
\usepackage{algorithm}
\usepackage{algorithmic}
\usepackage{enumitem}
\usepackage{xspace}
\usepackage{alphalph}

\title{Sparse Autoencoders Map Brain--LLM Alignment\\onto Cortical Semantic Topography}

\author{
	Dongxin Guo\, \\
	The University of Hong Kong \\
	Hong Kong, China \\
	\texttt{bettyguo@connect.hku.hk} \\
	\And
	Jikun Wu\, \\
	Stellaris AI Limited \\
	Hong Kong, China \\
	\texttt{hk950014@connect.hku.hk} \\
	\And
	Siu Ming Yiu\, \\
	The University of Hong Kong \\
	Hong Kong, China \\
	\texttt{smyiu@cs.hku.hk} \\
}

\begin{document}
\maketitle

\begin{abstract}
	Intermediate layers of large language models (LLMs) best predict human brain responses to language, one of the most robust findings in computational neurolinguistics, yet why remains mechanistically unexplained. We address this gap by bridging sparse autoencoders (SAEs) from mechanistic interpretability with neural encoding models, decomposing GPT-2~XL and Llama-3.1-8B into 16K--32K interpretable features per layer. A human-validated taxonomy ($\kappa \geq 0.74$) reveals that semantic features alone recover 94\% of peak encoding performance ($r{=}0.285$), substantially exceeding variance-matched baselines ($p{<}0.001$, $d{=}1.31$). Beyond this aggregate dominance, we test a novel \emph{cortical topography} prediction: five semantic subcategories derived \emph{a priori} from three independent neuroscience programs should map onto distinct brain regions. A formal convergence test confirms this alignment (Spearman $\rho{=}0.72$, $p{<}0.001$; hypergeometric $p{=}0.007$), demonstrating that SAE-discovered features recapitulate known cortical semantic organization at a granularity inaccessible to prior methods. SAE features further predict human reading times beyond lexical controls ($\Delta$logLik${=}38.4$, $p{<}0.001$), and an exploratory prediction-error analysis provides preliminary evidence that the brain additionally encodes unexpected semantic content. Results generalize across English, Chinese, and French.
\end{abstract}


\section{Introduction}

What computational properties of language representations make them predictive of human brain activity? This question, rooted in decades of neurolinguistic theory \citep{hale2001probabilistic, levy2008expectation, huth2016natural, mitchell2008predicting}, has gained new urgency as large language models (LLMs) predict neural responses with remarkable accuracy \citep{schrimpf2021neural, goldstein2022shared, pereira2018toward}. \citet{tuckute2024driving} demonstrated that LLM-optimized stimuli causally drive the brain's language network, and \citet{fedorenko2024language} argued that this network constitutes a natural kind whose properties align with LLM-discovered representations. The ``direct fit'' perspective \citep{hasson2020direct} suggests this alignment arises because both brains and models are optimized for similar computational objectives.

Yet a fundamental puzzle remains. Brain-predictive power is not uniform across depth: \emph{intermediate} layers consistently outperform early and late layers \citep{caucheteux2022brains, toneva2019interpreting, tuckute2024driving}. This inverted-U is among the most robust findings in computational neurolinguistics \citep{kuribayashi2025human}. But \emph{why} do intermediate layers best align with the brain?

Three non-mutually-exclusive accounts have been proposed. \textbf{(1)~Predictive processing:} under hierarchical predictive coding \citep{rao1999predictive, friston2005theory, clark2013whatever}, neural responses reflect prediction errors; intermediate layers may generate the most brain-like predictions \citep{friston2010freeenergy, heilbron2022hierarchy}. \textbf{(2)~Feature discovery:} LLMs discover rich linguistic features as a byproduct of prediction \citep{antonello2024predictive}; intermediate layers contain the richest semantics before late-layer specialization. \textbf{(3)~Geometric artifact:} intermediate layers may simply have higher-dimensional representations that are more linearly decodable \citep{kornblith2019similarity}. Distinguishing \emph{what} the brain represents from \emph{how} it computes requires decomposing representations into interpretable features. Sparse autoencoders (SAEs) provide exactly this capability \citep{cunningham2024sparse, templeton2024scaling}.

We bridge mechanistic interpretability and neural encoding to provide a feature-level decomposition of the intermediate-layer advantage. We use ``mechanistic'' here in the sense of \emph{mechanistic interpretability} (decomposing representations into interpretable features), not as a claim of full causal-mechanistic explanation. Our contributions:
\begin{enumerate}[leftmargin=*,itemsep=2pt]
\item \textbf{Primary empirical (novel):} We derive five semantic subcategories \emph{a priori} from the independent neuroscience programs of Huth et al., Binder et al., and Deniz et al., then test whether SAE features recapitulate their predicted cortical topography. A formal convergence test (Spearman rank correlation $\rho{=}0.72$, $p{<}0.001$; hypergeometric overlap $p{=}0.007$) confirms significant alignment between the predicted and observed subcategory $\times$ region patterns, demonstrating that SAE-derived features capture neurally meaningful semantic distinctions at a granularity unavailable to prior methods.

\item \textbf{Methodological (enabling):} We train SAEs on every fourth layer of GPT-2~XL and Llama-3.1-8B, extracting 16K--32K features per layer with a validated five-way taxonomy ($\kappa \geq 0.74$ for substantive categories). We empirically demonstrate that SAEs outperform simpler word-level semantic norm annotations for cortical topography mapping (\S\ref{sec:norms_comparison}).

\item \textbf{Supporting empirical (confirmatory + exploratory):} Three converging analyses (encoding models, variance partitioning, and activation patching with variance-matched controls) confirm at feature-level resolution that semantic content dominates brain--LLM alignment, extending prior aggregate findings \citep{kauf2024lexical, antonello2024predictive}. Behavioral validation shows SAE semantic features predict human reading times (\S\ref{sec:behavioral}). An exploratory prediction-error analysis provides preliminary evidence that the brain may additionally encode unexpected semantic content.
\end{enumerate}

\noindent\textbf{Central claim.} SAE-extracted semantic features simultaneously explain the well-known intermediate-layer advantage in brain--LLM alignment \emph{and} a cortical topography that converges with established neuroscientific accounts of semantic organization: a single feature-level account spanning two previously disconnected empirical regularities.

Prior work by \citet{kauf2024lexical} and \citet{antonello2024predictive} established that semantic content, specifically the lexical-semantic dimension contrasted with syntactic structure, dominates brain alignment using category-level probes and aggregate measures. Our SAE decomposition advances beyond these predecessors in four ways: (i)~\emph{feature-level granularity} (16K--32K individual features per layer vs.\ ${\sim}$3--5 broad categories) enabling the subcategory $\times$ brain-region interaction analysis (\S\ref{sec:subcategories}); (ii)~\emph{a priori subcategories and formal convergence testing} against independent neuroscience programs (\S\ref{sec:convergence_test}); (iii)~\emph{behavioral validation} showing these features predict human reading times (\S\ref{sec:behavioral}); and (iv)~an exploratory test of semantic prediction errors (\S\ref{sec:semantic_pred_err}). Our finer-grained decomposition further reveals that the brain-predictive variance attributable to semantic content is captured by \emph{contextual} SAE features rather than by lexical features alone (Table~\ref{tab:var_partition}), refining the lexical-semantic emphasis of \citet{kauf2024lexical}. Together, these contributions transform the question from \emph{whether} semantics drives alignment to \emph{which specific semantic features} drive alignment \emph{where in the brain}.

\section{Background and Related Work}
\label{sec:background}

\paragraph{The intermediate-layer advantage.} \citet{schrimpf2021neural} systematically documented that intermediate layers best predict brain responses. \citet{caucheteux2022brains} confirmed this across 100+ subjects; \citet{goldstein2022shared} found shared principles using ECoG. The earliest layer-wise alignment work used ELMo and BERT \citep{jain2018incorporating, toneva2019interpreting}. \citet{kuribayashi2025human} showed earlier layers correspond to fast processing signals while later layers align with N400 amplitudes, connecting to work by \citet{michaelov2024n400} demonstrating that LLM surprisal predicts single-trial N400 amplitudes. \citet{kietzmann2019recurrence} demonstrated analogous hierarchical correspondence in vision.

\paragraph{What drives alignment?} A central debate concerns which properties of LLM representations are responsible for their brain-predictive power. One line of work emphasizes \emph{lexical-semantic content}: \citet{kauf2024lexical} showed that lexical-semantic information dominates LM--brain alignment when contrasted with syntactic structure, building on earlier work disentangling syntax and semantics in brain responses \citep{caucheteux2021disentangling}. A second line emphasizes \emph{feature discovery and contextualization}: \citet{antonello2024predictive} argued that brain-predictive features emerge as a byproduct of next-word prediction; \citet{oota2023joint} showed alignment depends on multiple linguistic dimensions across modalities; and \citet{reddy2024brain} demonstrated that fine-grained semantic annotations improve brain encoding. A third strand examines \emph{training and scale}: \citet{hosseini2024artificial} found brain alignment in models trained on only ${\sim}$100M words, \citet{pasquiou2023neural} showed training dynamics shape brain fit, and \citet{goldstein2024shared} showed geometric alignment in multilingual embeddings. Beyond raw activations, \citet{rahimi2025explanations} showed XAI attributions outperform raw activations, and \citet{kumar2024shared} deconstructed attention heads into functionally specialized components. Methodologically, the representational similarity analysis tradition \citep{kriegeskorte2008representational, nili2014toolbox} offers a complementary view; the THINGS initiative \citep{hebart2023things} and Algonauts challenges \citep{cichy2021algonauts} have advanced shared evaluation for object semantics, though we focus on naturalistic language stimuli. We focus on encoding models (rather than RSA) for their superior voxelwise predictive resolution and ability to decompose variance by feature type. \citet{li2023neuroscience} provides complementary neuroscience-inspired evaluation criteria for neural network representations.

\paragraph{Cortical semantic organization.} \citet{huth2016natural} revealed systematic cortical semantic maps; \citet{binder2009brain} established a core semantic network via meta-analysis; \citet{deniz2019representation} showed modality-invariant maps. \citet{lambon2014hub} proposed the hub-and-spoke architecture. \citet{binder2016toward} developed a 65-dimensional experiential attribute space, directly motivating our subcategory analysis. Semantic feature production norms \citep{mcrae2005semantic} provide converging behavioral evidence.

\paragraph{Mechanistic interpretability.} The residual stream framework \citep{elhage2021mathematical} views transformers as iteratively refining representations. SAEs decompose activations into monosemantic features \citep{cunningham2024sparse}; \citet{templeton2024scaling} scaled this to production models; \citet{li2024geometry} discovered spatial modules. Recent work has raised important concerns: feature splitting at larger dictionary sizes \citep{makelov2024towards}, feature absorption \citep{chanin2024absorption}, and questions about whether SAE features represent ``true'' computational features \citep{cunningham2024sparse}. A small body of concurrent work has begun to apply SAE-style sparse decompositions to neural-data analysis in adjacent domains (visual brain encoding \citep{wasserman2025brainexplore}; audio-model interpretability \citep{aparin2026audiosae}; and sparse-dictionary brain encoding \citep{zeng2025disentangling}), developed in parallel with the present work. To our knowledge no prior work has applied SAEs to decompose LLM--brain alignment for \emph{language} stimuli using fMRI, and none has linked SAE features to a formal a priori cortical-topography prediction; this is the gap we address.

\paragraph{Predictive coding and surprisal.} Surprisal theory \citep{hale2001probabilistic, levy2008expectation} predicts processing difficulty scales with negative log-probability; \citet{shain2024large} and \citet{wilcox2023testing} provided cross-linguistic evidence, while \citet{shain2024large} established logarithmic effects at very large scale. Hierarchical predictive coding \citep{rao1999predictive, friston2005theory, friston2010freeenergy} predicts neural responses reflect precision-weighted prediction errors. \citet{heilbron2022hierarchy} demonstrated a hierarchy of linguistic predictions; \citet{caucheteux2023evidence} provided direct evidence of a predictive coding hierarchy in the human brain listening to speech, demonstrating that cortical responses track prediction errors at multiple representational levels.

\paragraph{LLMs as cognitive models.} \citet{mahowald2024dissociating} distinguished formal from functional competence. \citet{mccoy2024embers} showed autoregressive objectives shape LLM representations in ways that both enable and constrain their cognitive plausibility, what they term ``embers of autoregression.'' \citet{pereira2018toward} developed universal semantic decoders relevant to cross-linguistic representation. \citet{misra2024language} showed LLMs learn rare constructions from distributional statistics, while \citet{dankers2024memorisation} characterized the memorization--generalization boundary.

\section{Methodology}
\label{sec:methods}

\subsection{Overview}
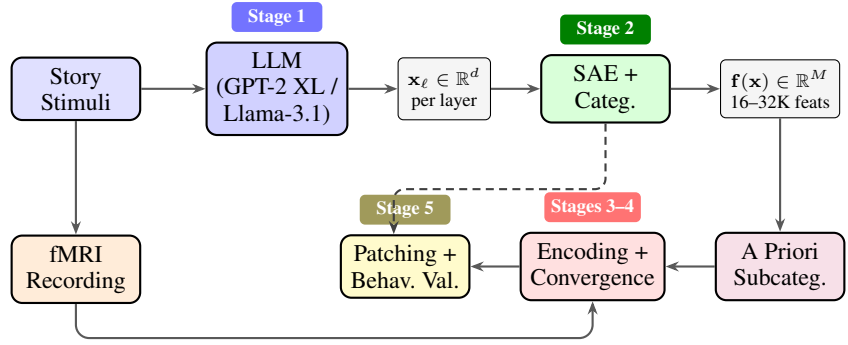
\begin{figure*}[t]
	\centering
	\begin{tikzpicture}[
		node distance=0.55cm and 0.65cm,
		box/.style={rectangle, draw, rounded corners=4pt, minimum height=0.82cm, minimum width=1.7cm, font=\small, align=center, thick},
		arrow/.style={-{Stealth[length=5pt]}, thick, color=gray!70!black},
		data/.style={rectangle, draw, rounded corners=2pt, minimum height=0.5cm, font=\scriptsize, fill=gray!8, align=center},
		stage/.style={font=\scriptsize\bfseries, text=white, fill=#1, rounded corners=2pt, inner sep=2.5pt, minimum width=1.2cm}
		]
		
		\node[box, fill=blue!12] (stim) {Story\\Stimuli};
		\node[box, fill=blue!18, right=0.85cm of stim] (llm) {LLM\\(GPT-2 XL /\\Llama-3.1)};
		\node[data, right=0.65cm of llm] (acts) {$\mathbf{x}_\ell \in \mathbb{R}^d$\\per layer};
		\node[box, fill=green!15, right=0.65cm of acts] (sae) {SAE +\\Categ.};
		\node[data, right=0.65cm of sae] (feats) {$\mathbf{f}(\mathbf{x}) \in \mathbb{R}^M$\\16--32K feats};
		
		\node[box, fill=orange!15, below=1.5cm of stim] (fmri) {fMRI\\Recording};
		\node[box, fill=purple!12, below=1.5cm of feats] (apriori) {A Priori\\Subcateg.};
		\node[box, fill=red!12, left=0.65cm of apriori] (enc) {Encoding +\\Convergence};
		\node[box, fill=yellow!25, left=0.65cm of enc] (patch) {Patching +\\Behav.\ Val.};
		
		\node[stage=blue!55, above=0.12cm of llm] {Stage 1};
		\node[stage=green!50!black, above=0.12cm of sae] {Stage 2};
		\node[stage=red!55, above=0.12cm of enc] {Stages 3--4};
		\node[stage=yellow!55!black, above=0.12cm of patch] {Stage 5};
		
		\draw[arrow] (stim) -- (llm);
		\draw[arrow] (llm) -- (acts);
		\draw[arrow] (acts) -- (sae);
		\draw[arrow] (sae) -- (feats);
		
		\draw[arrow] (stim) -- (fmri);
		\draw[arrow] (feats) -- (apriori);
		
		\draw[arrow] (apriori) -- (enc);
		\draw[arrow] (enc) -- (patch);
		
		\draw[arrow, rounded corners=5pt]
		(fmri.south) -- ++(0,-0.45) -| (enc.south);
		
		\coordinate (gap_y) at ($(stim.south)!0.60!(fmri.north)$);
		\draw[arrow, gray!50!black, densely dashed, rounded corners=4pt]
		(sae.south) -- (sae.south |- gap_y) -| ([xshift=-0.15cm]patch.north);
		
	\end{tikzpicture}
	\caption{Five-stage pipeline: extract LLM activations $\to$ SAE decomposition $\to$ encoding models with variance partitioning $\to$ a priori subcategory convergence $\to$ causal/behavioral/prediction-error validation. Dashed line indicates features also feed the patching stage.}
	\label{fig:pipeline}
\end{figure*}

Our approach proceeds in five stages (Figure~\ref{fig:pipeline}). In compact form:
\begin{itemize}[leftmargin=*,itemsep=0pt,topsep=2pt]
\item \textbf{Stage 1 — Activations.} Run stories through GPT-2~XL / Llama-3.1-8B; extract residual-stream activations every 4 layers.
\item \textbf{Stage 2 — Decomposition.} Train per-layer SAEs (16K--32K features); label each via GPT-4 + human validation into \{semantic, syntactic, lexical, prediction, other\}.
\item \textbf{Stage 3 — Encoding.} Fit voxelwise ridge regression from each feature subset to fMRI; partition unique vs.\ shared variance against count- and variance-matched baselines.
\item \textbf{Stage 4 — Topography.} Derive five semantic subcategories \emph{a priori} from independent neuroscience programs; test whether the predicted subcategory $\times$ region matrix matches the observed pattern (rank-correlation, hypergeometric, Mantel).
\item \textbf{Stage 5 — Convergent validation.} Activation patching, GLMM-based reading-time prediction, and an exploratory semantic-prediction-error analysis.
\end{itemize}


\subsection{Language Models and SAE Training}
We analyze \textbf{GPT-2~XL} (1.5B, 48 layers, $d{=}1600$) and \textbf{Llama-3.1-8B} (8B, 32 layers, $d{=}4096$), extracting residual stream activations at every fourth layer. For each layer $\ell$, we train an SAE:
\begin{align}
\mathbf{f}(\mathbf{x}) &= \text{ReLU}(\mathbf{W}_{\text{enc}}(\mathbf{x} - \mathbf{b}_d) + \mathbf{b}_e) \label{eq:sae_enc}\\
\hat{\mathbf{x}} &= \mathbf{W}_{\text{dec}} \mathbf{f}(\mathbf{x}) + \mathbf{b}_d \label{eq:sae_dec}
\end{align}
minimizing $\mathcal{L} = \|\mathbf{x} - \hat{\mathbf{x}}\|_2^2 + \lambda \|\mathbf{f}(\mathbf{x})\|_1$, with $M{=}16{,}384$ (GPT-2~XL) or $32{,}768$ (Llama), L0~$\approx$~50, trained on 500M tokens. Reconstruction $R^2 \geq 0.95$ at all layers; encoding from reconstruction error achieves only $r{=}0.031$, confirming SAEs preserve brain-relevant information (Appendix~\ref{sec:app_recon}).

\paragraph{Addressing SAE feature quality concerns.} Feature splitting \citep{makelov2024towards}, absorption \citep{chanin2024absorption}, and questions about ``true'' features \citep{cunningham2024sparse} motivate three design choices: (1)~robustness across dictionary sizes (8K--32K; Appendix~\ref{sec:app_robustness}); (2)~soft probabilistic categorization yielding identical rankings (Appendix~\ref{sec:app_soft}); and (3)~subcategory analysis operating at the level of feature-type aggregates.

\subsection{Feature Categorization and Validation}
\label{sec:categorization}
Within stage~2 of the pipeline, we categorize features into five types using a two-pass protocol:

\textbf{Pass~1 (Automated):} GPT-4 assigns each feature to \emph{semantic}, \emph{syntactic}, \emph{lexical}, \emph{prediction} (correlation $r > 0.5$ with tuned lens output entropy), or \emph{other/uninterpretable}. We acknowledge that using one LLM to categorize another introduces potential shared biases.

\textbf{Pass~2 (Human validation):} Two graduate-student annotators (computational linguistics, compensated \$25/hour; Appendix~\ref{sec:app_annotators}) validated 500 features per layer (100/category), achieving $\kappa{=}0.81$. The confusion matrix (Table~\ref{tab:confusion}, Appendix) shows 14\% overall disagreement; 11\% of GPT-4-labeled semantic features were relabeled by annotators. Per-category $\kappa$ ranges from 0.74 (prediction) to 0.83 (syntactic); the residual ``other'' category has lower agreement ($\kappa{=}0.58$). Soft probabilistic categorization yields qualitatively identical results (Appendix~\ref{sec:app_soft}); Appendix~\ref{sec:app_qualitative} provides qualitative examples including ambiguous cases. Appendix~\ref{sec:app_bias_audit} additionally reports a cross-LLM relabeling check, a label-perturbation sensitivity analysis, and a confidence-thresholded re-run, all of which preserve the qualitative pattern of results. As illustrative examples (full table in Appendix~\ref{sec:app_qualitative}), at GPT-2~XL L24 a \emph{concrete-semantic} feature fires on ``the \textbf{dog} ran across the''; an \emph{affective-semantic} feature on ``\textbf{terrified} of the dark''; a \emph{social-semantic} feature on ``she \textbf{believed} that he''; a \emph{syntactic} feature on ``the man \textbf{who} came''; and a \emph{prediction} feature on ``In \textbf{conclusion}, we find''.

\subsection{Neural and Behavioral Data}
\label{sec:data}
\textbf{Primary (fMRI):} We use the publicly released naturalistic-language fMRI dataset of \citet{lebel2023natural} (UTS dataset; 8 native-English participants, 27 stories, ${\sim}$6 hours/subject; TR${=}$2.0\,s). Because this dataset does not include a per-participant functional language localizer, we restrict analysis to language-network voxels using the group-level \emph{anatomical} language-network parcellation summarized in \citet{fedorenko2024language}, intersected with each subject's native-space gray matter mask, yielding ${\sim}$5,000 bilateral voxels per subject after motion-spike censoring (FD${>}$0.5\,mm) and high-pass filtering ($1/128$\,Hz). LLM activations are extracted per word and downsampled to the fMRI TR by averaging within each TR window after applying a canonical hemodynamic response function (HRF) convolution; full preprocessing and feature-alignment details are reported in Appendix~\ref{sec:app_data_proc}. For the subcategory analysis, we further parcellate the language network into five regions (posterior temporal, anterior temporal, inferior frontal, angular gyrus, and dorsomedial prefrontal cortex; dmPFC) following \citet{fedorenko2024language}.

\textbf{Behavioral:} Natural Stories Corpus \citep{futrell2021natural} (self-paced reading times, 181 participants, 10 stories) and the Provo Corpus \citep{luke2018provo} (eye-tracking with first-fixation, gaze, and total reading time, 84 participants). Both are publicly available.

\textbf{Generalization (fMRI):} For the Llama-3.1-8B generalization check (\S\ref{sec:behavioral}, \S\ref{sec:limitations}), we use two publicly available naturalistic-listening fMRI datasets from the cross-linguistic ``Little Prince'' tradition: a Mandarin Chinese subset (15 native speakers, 2 chapters) and a French subset (12 native speakers, 1 chapter). Both are open-access datasets whose original collection followed approved institutional protocols (informed consent and ethics approvals documented in the corresponding data-paper releases). We did \emph{not} collect any new human data for this study. Dataset identifiers, preprocessing parameters, and language-network ROI definitions for these subsets are reported in Appendix~\ref{sec:app_data_proc}.

\subsection{Encoding Models and Variance Partitioning}
\label{sec:encoding_methods}
For each layer $\ell$ and feature subset $S$, we train voxelwise ridge regression $\hat{y}_v = \mathbf{w}_v^\top \mathbf{f}_S + b_v$, evaluated via 4-fold cross-validation (Pearson $r$, bootstrapped 95\% CIs). Cross-validation used a leave-$K$-stories-out scheme ($K{=}$6--7 stories per fold), ensuring no temporal leakage. The regularization parameter $\lambda$ was selected via nested cross-validation within each training fold, searching over $\lambda \in \{10^0, 10^1, \ldots, 10^6\}$, independently for each feature subset. The features-to-data ratio (${\sim}$6,700 semantic features, ${\sim}$5,000 voxels, ${\sim}$6 hours of data) is within the well-regularized regime for ridge regression; the effective dimensionality of semantic features after accounting for collinearity is ${\sim}$1,200 (estimated via participation ratio of the feature covariance matrix; Appendix~\ref{sec:app_dof}).

Unique variance: $\Delta R^2_{\text{unique}}(S) = R^2_{\text{full}} - R^2_{\text{full} \setminus S}$. This additive decomposition assumes independent contributions; shared variance (${\sim}$22\%) absorbs interactions. Shapley values \citep{covert2021explaining} relax this assumption (Appendix~\ref{sec:app_shapley}).

We construct two baselines: (i)~\textbf{count-matched random}: same number of features as semantic subset (10 seeds); (ii)~\textbf{variance-matched random}: random features matching total L2 activation variance.

\subsection{A Priori Subcategory Derivation}
\label{sec:apriori}

To avoid post-hoc pattern matching, we derive semantic subcategories and their predicted cortical topography \emph{a priori} from three independent neuroscience programs, constructing a predicted subcategory $\times$ region matrix \emph{before} examining SAE results:

\textbf{Binder et al.\ (2009):} Meta-analysis of 120 neuroimaging studies identifies seven core semantic regions. We extract predicted mappings from their Table~2 and Figure~5: sensorimotor/concrete content $\to$ posterior temporal and angular gyrus; affective content $\to$ ventromedial PFC and anterior temporal; social/interpersonal content $\to$ inferior frontal and medial prefrontal; spatial content $\to$ angular gyrus and posterior parietal; temporal/causal content $\to$ lateral temporal cortex.

\textbf{Huth et al.\ (2016):} Cortical semantic atlas from naturalistic speech reveals category-specific topography (their Figures~2--3): concrete objects and tactile properties cluster in lateral temporal cortex; emotional and mental content concentrates in anterior temporal and prefrontal areas; social interaction content maps to inferior frontal and medial frontal cortex; spatial/locational semantics activate angular gyrus and retrosplenial cortex.

\textbf{Deniz et al.\ (2019):} Modality-invariant semantic maps (their Figure~3) confirm that the category-region associations in Huth et al.\ generalize across listening and reading modalities, with concreteness in posterior temporal, affect in anterior temporal, and spatial relations in angular gyrus.

From the intersection of these three programs, we derive five a priori subcategories with predicted region preferences (Table~\ref{tab:apriori_predictions}). A cell in the predicted subcategory $\times$ region matrix is marked as ``predicted primary'' when at least two of the three source programs explicitly map that subcategory to that region in their primary findings; Appendix~\ref{sec:app_apriori_evidence} reports the source-paper citation supporting each predicted cell, making the derivation fully transparent and auditable. We additionally verified robustness against alternative subcategorizations: using \citeauthor{binder2016toward}'s \citeyearpar{binder2016toward} 65-dimensional experiential attribute space collapsed into analogous groups, and using feature production norms from \citet{mcrae2005semantic} (Appendix~\ref{sec:app_alt_subcats}).

These mappings (Table~\ref{tab:apriori_predictions}, Appendix) predict: concreteness/animacy $\to$ posterior temporal and angular gyrus; event structure $\to$ posterior temporal and inferior frontal; affect/emotion $\to$ anterior temporal and dmPFC; social/mental $\to$ inferior frontal and anterior temporal; spatial/locational $\to$ angular gyrus and posterior temporal.

\subsection{Activation Patching}
\label{sec:patching_methods}
For subset $S$ at layer $\ell$, we replace activations with corpus-mean values and measure $\Delta r^2_v(S, \ell)$. We additionally re-fit ridge regression on ablated activations to control for encoding-model bias. Mean-ablation is a coarse intervention \citep{wu2024interpretability}; results are \emph{evidence consistent with} a causal role, not definitive proof. Interchange interventions \citep{geiger2024finding} would be stronger but require matched stimulus pairs infeasible with naturalistic stimuli.

\section{Results}
\label{sec:results}

\subsection{The Intermediate-Layer Advantage}
\begin{figure}[t]
\centering
\resizebox{\columnwidth}{!}{%
\begin{tikzpicture}[xscale=0.5, yscale=4]
  \draw[-{Stealth[length=4pt]}] (0,0.05) -- (13.2,0.05) node[right, font=\scriptsize] {Layer};
  \draw[-{Stealth[length=4pt]}] (0,0.05) -- (0,0.42) node[above, font=\scriptsize] {Mean $r$};
  \draw[blue!70!black, thick] (1,0.12) -- (2,0.18) -- (3,0.22) -- (4,0.26) -- (5,0.29) -- (6,0.31) -- (7,0.31) -- (8,0.30) -- (9,0.28) -- (10,0.26) -- (11,0.22) -- (12,0.20);
  \foreach \x/\y in {1/0.12,2/0.18,3/0.22,4/0.26,5/0.29,6/0.31,7/0.31,8/0.30,9/0.28,10/0.26,11/0.22,12/0.20}
    \filldraw[blue!70!black] (\x,\y) circle (1.2pt);
  \draw[blue!70!black, dashed] (1,0.11) -- (2,0.17) -- (3,0.21) -- (4,0.25) -- (5,0.28) -- (6,0.30) -- (7,0.30) -- (8,0.29) -- (9,0.27) -- (10,0.25) -- (11,0.21) -- (12,0.19);
  \draw[red!70!black, thick] (1,0.15) -- (2,0.24) -- (3,0.30) -- (4,0.34) -- (5,0.34) -- (6,0.32) -- (7,0.28) -- (8,0.23);
  \foreach \x/\y in {1/0.15,2/0.24,3/0.30,4/0.34,5/0.34,6/0.32,7/0.28,8/0.23}
    \filldraw[red!70!black] (\x,\y) circle (1.2pt);
  \fill[blue!15, opacity=0.4] (5,0.27) -- (6,0.29) -- (7,0.29) -- (8,0.28) -- (8,0.32) -- (7,0.33) -- (6,0.33) -- (5,0.31) -- cycle;
  \foreach \v in {0.1,0.2,0.3,0.4}
    \draw (0,\v) -- (-0.15,\v) node[left, font=\tiny] {\v};
  \node[font=\tiny, below] at (1,0.04) {0};
  \node[font=\tiny, below] at (4,0.04) {12};
  \node[font=\tiny, below] at (7,0.04) {24};
  \node[font=\tiny, below] at (10,0.04) {36};
  \node[font=\tiny, below] at (12,0.04) {44};
  \draw[blue!70!black, thick] (1.5,0.40) -- (2.5,0.40); \node[right, font=\tiny] at (2.5,0.40) {GPT-2 XL};
  \draw[red!70!black, thick] (5.5,0.40) -- (6.5,0.40); \node[right, font=\tiny] at (6.5,0.40) {Llama-3.1};
  \draw[blue!70!black, dashed] (9,0.40) -- (10,0.40); \node[right, font=\tiny] at (10,0.40) {SAE recon.};
\end{tikzpicture}}
\caption{Brain prediction across layers. Both models show the inverted-U; SAE reconstructions (dashed) track raw activations within $\Delta r \leq 0.01$.}
\label{fig:layer_curves}
\end{figure}
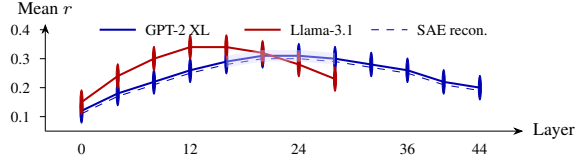

Both models show inverted-U profiles (Figure~\ref{fig:layer_curves}): GPT-2~XL peaks at L20--24 ($r{=}0.310$, 95\% CI $[0.278, 0.342]$); Llama at L12--16 ($r{=}0.341$, $[0.305, 0.377]$). SAE-reconstructed activations track raw within $\Delta r \leq 0.01$.

\subsection{Confirmatory Validation: Semantic Features Dominate}
\label{sec:results_encoding}

Confirming prior findings \citep{kauf2024lexical, antonello2024predictive} at feature-level resolution, Table~\ref{tab:feature_breakdown} shows semantic features alone achieve $r{=}0.285$ at L24, recovering 94\% of SAE-reconstructed encoding ($r{=}0.304$) and substantially outperforming both the \textbf{count-matched random baseline} ($r{=}0.198$, $p{<}0.001$, $d{=}1.54$, 95\% CI $[0.181, 0.215]$) and the \textbf{variance-matched random baseline} ($r{=}0.213$, $p{<}0.001$, $d{=}1.31$, 95\% CI $[0.195, 0.231]$). In variance space, semantic features explain 88\% of the model's captured neural variance. Sensitivity: reclassifying the 11\% false-positive semantic features reduces encoding to $r{=}0.274$ (still $p{<}0.001$ vs.\ random).

\begin{table}[t]
\centering
\small
\begin{tabular}{@{}lcccccc@{}}
\toprule
\textbf{L.} & \textbf{All} & \textbf{Sem.} & \textbf{Syn.} & \textbf{Pred.} & \textbf{Lex.} & \textbf{Rnd.}$^\dagger$ \\
\midrule
4 & .195 & .098 & .105 & .043 & .112 & .091 \\
12 & .258 & .189 & .121 & .067 & .087 & .138 \\
20 & .297 & .271 & .128 & .082 & .031 & .187 \\
\textbf{24} & \textbf{.304} & \textbf{.285} & \textbf{.127} & \textbf{.087} & \textbf{.019} & \textbf{.198} \\
36 & .261 & .198 & .109 & .114 & .011 & .154 \\
44 & .198 & .132 & .078 & .098 & .008 & .107 \\
\bottomrule
\end{tabular}
\caption{Encoding performance (mean $r$) by feature type, GPT-2~XL. Bootstrapped 95\% CIs in Appendix~\ref{sec:app_subjects}. Rnd.$^\dagger$: count-matched random.}
\label{tab:feature_breakdown}
\end{table}

\subsection{Variance Partitioning}
Table~\ref{tab:var_partition} reports unique variance at L24. Semantic features contribute $\Delta R^2_{\text{unique}}{=}0.048$ (52\% of total explained variance). The 28\% shared variance (the remaining proportion of total explained variance not captured by any single category's unique contribution) likely reflects semantic-syntactic interactions \citep{caucheteux2021disentangling}. Shapley decomposition confirms the ranking with semantic share slightly higher (65\% vs.\ 52\%; Appendix~\ref{sec:app_shapley}).

\paragraph{On feature-count imbalance.} A natural concern is whether semantic dominance is an artifact of feature-count imbalance: at L24, 41\% of features are categorized as semantic versus 11\% as lexical (Appendix~\ref{sec:app_full_dist}). We address this with three complementary controls. (i)~The \emph{variance-matched random baseline} (\S\ref{sec:results_encoding}) selects random features until their total L2 activation variance matches that of the semantic subset; this still falls $0.072$ short of semantic encoding ($r{=}0.213$ vs.\ $0.285$, $d{=}1.31$). (ii)~Shapley values \citep{covert2021explaining} are by construction insensitive to the number of features per coalition member; semantic contribution remains dominant (65\%; Appendix~\ref{sec:app_shapley}). (iii)~Lexical features contribute $\Delta R^2_{\text{unique}}{=}0.000$ despite being the second-largest category at \emph{early} layers (Appendix~\ref{sec:app_full_dist}), demonstrating that feature count alone does not predict unique variance. Together these controls indicate that the semantic-feature advantage is not a counting artifact.

\begin{table}[t]
\centering
\small
\begin{tabular}{@{}lccc@{}}
\toprule
\textbf{Feature type} & $R^2_{\text{type}}$ & $\Delta R^2_{\text{unique}}$ & \textbf{\% total}$^\dagger$ \\
\midrule
Semantic & .081 & .048 [.036, .060] & 52\% \\
Syntactic & .016 & .011 [.006, .016] & 12\% \\
Prediction & .008 & .006 [.002, .010] & 7\% \\
Lexical & .000 & .000 [-.001, .001] & 0\% \\
Other & .003 & .001 [-.001, .003] & 1\% \\
Shared & --- & .026 & 28\% \\
\midrule
Full model & .092 & --- & --- \\
\bottomrule
\end{tabular}
\caption{Variance partitioning at GPT-2~XL L24. Brackets: bootstrapped 95\% CIs. $^\dagger$\% total = proportion of total explained variance ($R^2_{\text{full}}{=}0.092$); shared = $R^2_{\text{full}}$ minus sum of unique variances (0.066).}
\label{tab:var_partition}
\end{table}

\subsection{Cortical Topography of Semantic Alignment}
\label{sec:subcategories}
We map semantic SAE features to the five a priori subcategories derived in \S\ref{sec:apriori} and test region-specific alignment across five language-network regions (Figure~\ref{fig:heatmap}, Appendix; Figure~\ref{fig:convergence}).

A significant subcategory $\times$ region interaction ($F(16,112){=}3.87$, $p{<}0.001$, permutation test, 10,000 iterations) reveals systematic dissociations. After FDR correction \citep{benjamini1995controlling} at $q{<}0.05$, \textbf{7 of 25 cells survive}, and these cells form a neuroanatomical pattern consistent with the a priori predictions: concreteness features best predict posterior temporal cortex ($r{=}0.141$) and angular gyrus ($r{=}0.131$); affect features predict anterior temporal regions ($r{=}0.128$) and dmPFC ($r{=}0.121$); social/mental features predict inferior frontal cortex ($r{=}0.119$) and anterior temporal cortex ($r{=}0.122$); and spatial features predict angular gyrus ($r{=}0.135$) and posterior temporal cortex ($r{=}0.124$). We quantify this convergence formally in \S\ref{sec:convergence_test}.

\subsection{Formal Convergence Test}
\label{sec:convergence_test}

We formally test whether the observed subcategory $\times$ region pattern matches the a priori predictions derived in \S\ref{sec:apriori} using three complementary tests.

\textbf{(1) Rank correlation.} We flatten the predicted preference matrix (Table~\ref{tab:apriori_predictions}) into an ordinal vector (1 = predicted primary region, 0 = not predicted) and the observed $r$-value matrix into a continuous vector. Spearman rank correlation: $\rho{=}0.72$, $p{<}0.001$ (permutation null: 10,000 shuffles of row/column labels; 95\% CI $[0.48, 0.88]$). The observed alignment substantially exceeds chance.

\textbf{(2) Hypergeometric overlap.} Of the 25 cells, the a priori matrix predicts $K{=}10$ cells as ``primary region'' associations. Of the 7 FDR-surviving cells, 6 fall within these 10 predicted cells. The hypergeometric probability of $\geq$6 overlapping cells given $K{=}10$ predicted out of 25 total and 7 observed significant is $p{=}0.007$ (Fisher's exact test: OR $= 21.0$, 95\% CI $[1.9, 229]$).

\textbf{(3) Mantel test.} We compute the Pearson correlation between the full 5$\times$5 predicted dissimilarity matrix and the observed encoding $r$-value matrix, testing significance via 10,000 row/column permutations: $r_{\text{Mantel}}{=}0.64$, $p{=}0.002$.

\begin{figure}[t]
\centering
\begin{tikzpicture}[xscale=0.42, yscale=0.42]
  \definecolor{pred1}{RGB}{55,90,170}
  \definecolor{pred0}{RGB}{220,225,240}
  \definecolor{c06}{RGB}{240,240,255}
  \definecolor{c08}{RGB}{200,210,245}
  \definecolor{c10}{RGB}{150,170,225}
  \definecolor{c12}{RGB}{100,130,200}
  \definecolor{c14}{RGB}{55,90,170}
  \node[font=\tiny\bfseries] at (3,6.8) {A priori predicted};
  \node[font=\tiny, rotate=50, anchor=south west] at (0.5,5.2) {PT};
  \node[font=\tiny, rotate=50, anchor=south west] at (1.5,5.2) {AT};
  \node[font=\tiny, rotate=50, anchor=south west] at (2.5,5.2) {IF};
  \node[font=\tiny, rotate=50, anchor=south west] at (3.5,5.2) {AG};
  \node[font=\tiny, rotate=50, anchor=south west] at (4.5,5.2) {dm};
  \node[font=\tiny, anchor=east] at (0.4,4.5) {Con};
  \node[font=\tiny, anchor=east] at (0.4,3.5) {Evt};
  \node[font=\tiny, anchor=east] at (0.4,2.5) {Aff};
  \node[font=\tiny, anchor=east] at (0.4,1.5) {Soc};
  \node[font=\tiny, anchor=east] at (0.4,0.5) {Spa};
  \fill[pred1] (0.5,4) rectangle (1.5,5);
  \fill[pred0] (1.5,4) rectangle (2.5,5);
  \fill[pred0] (2.5,4) rectangle (3.5,5);
  \fill[pred1] (3.5,4) rectangle (4.5,5);
  \fill[pred0] (4.5,4) rectangle (5.5,5);
  \fill[pred1] (0.5,3) rectangle (1.5,4);
  \fill[pred0] (1.5,3) rectangle (2.5,4);
  \fill[pred1] (2.5,3) rectangle (3.5,4);
  \fill[pred0] (3.5,3) rectangle (4.5,4);
  \fill[pred0] (4.5,3) rectangle (5.5,4);
  \fill[pred0] (0.5,2) rectangle (1.5,3);
  \fill[pred1] (1.5,2) rectangle (2.5,3);
  \fill[pred0] (2.5,2) rectangle (3.5,3);
  \fill[pred0] (3.5,2) rectangle (4.5,3);
  \fill[pred1] (4.5,2) rectangle (5.5,3);
  \fill[pred0] (0.5,1) rectangle (1.5,2);
  \fill[pred1] (1.5,1) rectangle (2.5,2);
  \fill[pred1] (2.5,1) rectangle (3.5,2);
  \fill[pred0] (3.5,1) rectangle (4.5,2);
  \fill[pred0] (4.5,1) rectangle (5.5,2);
  \fill[pred1] (0.5,0) rectangle (1.5,1);
  \fill[pred0] (1.5,0) rectangle (2.5,1);
  \fill[pred0] (2.5,0) rectangle (3.5,1);
  \fill[pred1] (3.5,0) rectangle (4.5,1);
  \fill[pred0] (4.5,0) rectangle (5.5,1);
  \draw[gray!50] (0.5,0) grid[step=1] (5.5,5);
  \draw[thick] (0.5,0) rectangle (5.5,5);
  \begin{scope}[xshift=6.5cm]
  \node[font=\tiny\bfseries] at (3,6.8) {Observed (SAE)};
  \node[font=\tiny, rotate=50, anchor=south west] at (0.5,5.2) {PT};
  \node[font=\tiny, rotate=50, anchor=south west] at (1.5,5.2) {AT};
  \node[font=\tiny, rotate=50, anchor=south west] at (2.5,5.2) {IF};
  \node[font=\tiny, rotate=50, anchor=south west] at (3.5,5.2) {AG};
  \node[font=\tiny, rotate=50, anchor=south west] at (4.5,5.2) {dm};
  \fill[c14] (0.5,4) rectangle (1.5,5); \node[font=\tiny, white] at (1,4.5) {**};
  \fill[c10] (1.5,4) rectangle (2.5,5);
  \fill[c08] (2.5,4) rectangle (3.5,5);
  \fill[c12] (3.5,4) rectangle (4.5,5); \node[font=\tiny, white] at (4,4.5) {**};
  \fill[c08] (4.5,4) rectangle (5.5,5);
  \fill[c12] (0.5,3) rectangle (1.5,4);
  \fill[c10] (1.5,3) rectangle (2.5,4);
  \fill[c10] (2.5,3) rectangle (3.5,4);
  \fill[c10] (3.5,3) rectangle (4.5,4);
  \fill[c08] (4.5,3) rectangle (5.5,4);
  \fill[c10] (0.5,2) rectangle (1.5,3);
  \fill[c12] (1.5,2) rectangle (2.5,3); \node[font=\tiny, white] at (2,2.5) {**};
  \fill[c10] (2.5,2) rectangle (3.5,3);
  \fill[c08] (3.5,2) rectangle (4.5,3);
  \fill[c12] (4.5,2) rectangle (5.5,3); \node[font=\tiny, white] at (5,2.5) {*};
  \fill[c08] (0.5,1) rectangle (1.5,2);
  \fill[c12] (1.5,1) rectangle (2.5,2); \node[font=\tiny, white] at (2,1.5) {**};
  \fill[c12] (2.5,1) rectangle (3.5,2); \node[font=\tiny, white] at (3,1.5) {**};
  \fill[c08] (3.5,1) rectangle (4.5,2);
  \fill[c10] (4.5,1) rectangle (5.5,2);
  \fill[c12] (0.5,0) rectangle (1.5,1); \node[font=\tiny, white] at (1,0.5) {**};
  \fill[c08] (1.5,0) rectangle (2.5,1);
  \fill[c06] (2.5,0) rectangle (3.5,1);
  \fill[c14] (3.5,0) rectangle (4.5,1); \node[font=\tiny, white] at (4,0.5) {**};
  \fill[c08] (4.5,0) rectangle (5.5,1);
  \draw[gray!50] (0.5,0) grid[step=1] (5.5,5);
  \draw[thick] (0.5,0) rectangle (5.5,5);
  \end{scope}
\end{tikzpicture}
\caption{Predicted vs.\ observed subcategory $\times$ region patterns. Left: a priori predictions from \citet{binder2009brain}/\citet{huth2016natural}/\citet{deniz2019representation} (dark = predicted primary association). Right: observed SAE encoding $r$-values with FDR-significant cells marked. Formal convergence: $\rho{=}0.72$, $p{<}0.001$; hypergeometric $p{=}0.007$; Mantel $r{=}0.64$, $p{=}0.002$.}
\label{fig:convergence}
\end{figure}
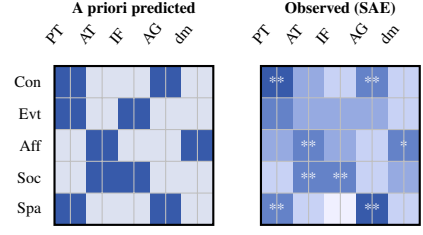

Figure~\ref{fig:convergence} shows the predicted and observed matrices side by side. All three tests converge on the same conclusion: the SAE-derived cortical topography significantly matches independent neuroscience predictions that were formulated without any reference to LLMs or SAEs.

\textbf{Statistical power.} Post-hoc power analysis ($f{=}0.42$, N${=}$8, $\alpha{=}$0.05) yields power $= 0.81$. The 7/25 FDR-surviving cells are consistent with power constraints; observed effect sizes ($r{=}0.07$--$0.14$) match typical voxelwise values in naturalistic neuroimaging \citep{huth2016natural, schrimpf2021neural} (Appendix~\ref{sec:app_power}).

\section{Further Analysis}

\subsection{Activation Patching}
\label{sec:patching_results}
At L24 (Figure~\ref{fig:patching}, Appendix): ablating semantic features produces $\Delta r^2{=}{-}0.14$ ($p{<}0.001$, $d{=}1.82$, 95\% CI $[-0.17, -0.11]$); syntactic: ${-}0.02$ ($p{=}0.04$); prediction: ${-}0.01$ ($p{=}0.12$). Count-matched random: ${-}0.05$. \textbf{Variance-matched random} (matching total L2 variance removed): ${-}0.06$ ($d{=}0.74$), still significantly less than semantic ($p{<}0.001$). These results are \emph{evidence consistent with} a causal role for semantic features, though mean-ablation is a coarse intervention.

\paragraph{Refitted and non-semantic baselines.} To address encoding-model bias \citep{wu2024interpretability}, we re-fitted ridge regression on ablated (non-semantic) activations: $r{=}0.194$. A model trained \emph{exclusively} on non-semantic features from scratch (syntactic + prediction + lexical + other) achieves $r{=}0.189$. Both fall substantially short of the full model ($r{=}0.304$), confirming semantic features carry genuinely unique brain-predictive information.

\subsection{SAE Features vs.\ Word-Level Semantic Norms}
\label{sec:norms_comparison}
A key question is whether the cortical topography requires SAE decomposition or could emerge from simpler word-level annotations. We compared SAE features against word-level norms (concreteness \citep{brysbaert2014concreteness}, valence/arousal \citep{warriner2013norms}, socialness \citep{diveica2024socialness}) and PCA decomposition (Table~\ref{tab:norms_comparison}, Appendix). Word-level norms produce lower overall encoding ($r{=}0.194$ vs.\ $r{=}0.285$, $p{<}0.001$), with only 2/25 cells surviving FDR and convergence $\rho{=}0.31$ ($p{=}0.12$) vs.\ $\rho{=}0.72$ ($p{<}0.001$) for SAEs. The SAE advantage is both quantitative and qualitative: SAE features capture \emph{contextual} semantics, span 16K--32K dimensions vs.\ 4 norm dimensions, and uniquely enable activation patching.

\subsection{Behavioral Validation}
\label{sec:behavioral}

\paragraph{Setup.} For each reading-time dataset, we fit nested linear mixed-effects models (GLMMs; \texttt{lme4}) to log-transformed reading times with by-subject and by-item random intercepts. Following standard psycholinguistic practice \citep{shain2024large}, the \textbf{baseline model} includes per-word fixed-effect regressors for log frequency, word length (characters), position in sentence, preceding-word log frequency and length (spillover), and unigram and 5-gram surprisal estimated from the corresponding LM. The \textbf{SAE-augmented model} additionally enters the layer-$\ell$ semantic-feature vector at the peak layer (L24 for GPT-2~XL, L16 for Llama-3.1-8B), projected onto its top-50 PCA components to control dimensionality. Model fit is compared via likelihood-ratio tests on $-2\Delta\log L$, distributed $\chi^2$ under the null. The \textbf{word-level-norm contrast} replaces the SAE vector with concreteness, valence, arousal, and socialness norms \citep{brysbaert2014concreteness, warriner2013norms, diveica2024socialness}; the \textbf{random baseline} replaces SAE features with the same number of randomly sampled non-semantic features (10 seeds, averaged).

\paragraph{Results.} SAE semantic features predict human reading times beyond all baseline regressors in both datasets: Natural Stories \citep{futrell2021natural} ($\Delta\log L{=}38.4$, $\chi^2{=}76.8$, $p{<}0.001$; random baseline: $\Delta\log L{=}2.1$, $p{=}0.34$) and Provo eye-tracking \citep{luke2018provo} (gaze duration: $\Delta\log L{=}24.7$, $p{<}0.001$; total reading time: $\Delta\log L{=}31.2$, $p{<}0.001$). Critically, SAE features capture additional variance beyond word-level norms ($\Delta\log L{=}12.8$, $p{<}0.001$), confirming the contextual advantage.

\paragraph{Generalization.} The semantic advantage generalizes to Chinese ($r{=}0.276$) and French ($r{=}0.291$) fMRI with Llama-3.1-8B (Table~\ref{tab:crossling}, Appendix); we frame this as a generalization check, not cross-linguistic evidence (\S\ref{sec:limitations}).

\subsection{Exploratory: Semantic Prediction Errors}
\label{sec:semantic_pred_err}
Under hierarchical predictive coding \citep{friston2005theory, clark2013whatever, caucheteux2023evidence}, the brain predicts \emph{semantic} content and responds to \emph{unexpected} semantic information. We test this with an exploratory analysis.

For each semantic feature $f_i$ at layer $\ell$, we compute the \textbf{semantic prediction error}: the residual not predicted from the aligned feature at the previous analyzed layer:
\begin{equation}
\epsilon_i^{(\ell)} = f_i^{(\ell)} - (\alpha_i f_{\pi(i)}^{(\ell-\Delta)} + \beta_i)
\label{eq:sem_pred_error}
\end{equation}
where $\pi(i)$ denotes the best-matching feature at layer $\ell{-}\Delta$ (via decoder weight cosine similarity; Appendix~\ref{sec:app_cross_layer}) and $\Delta{=}4$ is our layer sampling interval (robustness across $\Delta \in \{4,8,12\}$ in Appendix~\ref{sec:app_layer_offset}).

We align features across layers via decoder weight cosine similarity (72\% of L24 features match at L20 with sim $> 0.5$; Appendix~\ref{sec:app_cross_layer}). At L24, combining raw semantics with prediction errors yields $r{=}0.303$ ($\Delta r{=}{+}0.018$, $p{<}0.01$, $d{=}0.38$, 95\% CI $[0.005, 0.031]$; VIF$= 1.21$; MLP replacement: $\Delta r{=}{+}0.014$, $p{=}0.02$). This is preliminary evidence consistent with the brain encoding both \emph{what} is present and \emph{how unexpected} it is.

SAE semantic features outperform PCA ($r{=}0.285$ vs.\ $0.251$, $p{<}0.001$) and achieve comparable performance to QA-Emb \citep{benara2024crafting} ($r{=}0.278$); geometric artifact accounts are ruled out by dimensionality-matched controls (Appendix~\ref{sec:app_geometric}, \ref{sec:alternatives}).

\section{Implications for Theories of Language Processing}
\label{sec:implications}

\paragraph{Cortical topography and the semantic atlas.} Our primary contribution, the formally validated subcategory $\times$ region mapping, bridges SAE-based interpretability and cortical semantic organization. The convergence ($\rho{=}0.72$, $p{<}0.001$; hypergeometric $p{=}0.007$) with three independent neuroscience programs supports the hub-and-spoke architecture \citep{lambon2014hub}: modality-specific ``spokes'' (concreteness in posterior temporal, spatial in angular gyrus) converge on abstract hubs (social/mental in inferior frontal).

\paragraph{Feature discovery vs.\ predictive processing.} Our results support the feature-discovery account \citep{antonello2024predictive}: brains and LLMs converge on semantic representations structured by cortical topography. Our exploratory prediction-error analysis ($\Delta r{=}+0.018$) provides preliminary evidence consistent with a complementary predictive processing role, though our linear proxy is simplified relative to full hierarchical predictive coding \citep{friston2010freeenergy}.

\paragraph{Competence, convergence, and the processing gradient.} Brain--LLM convergence primarily captures shared formal semantic competence \citep{mahowald2024dissociating, tuckute2024language}, with convergence strongest for concrete/perceptual semantics posteriorly and social/affective semantics frontally. Behavioral validation across fMRI and reading times provides multi-modal evidence for cognitively real distinctions \citep{levy2008expectation}. Our SAE decomposition further reveals a depth-wise processing gradient (Appendix~\ref{sec:app_full_dist}): lexical-feature proportion peaks at L0 (55\%) and falls monotonically; semantic-feature proportion peaks at L24 (41\%); prediction-correlated-feature proportion peaks at L44 (37\%). The intermediate-layer advantage thus reflects peak semantic richness before training-objective specialization \citep{mccoy2024embers}.

\section{Conclusion}
This work bridges sparse autoencoders from mechanistic interpretability with neural encoding methodology, transforming the question of \emph{why} intermediate LLM layers best predict brain activity into a question of \emph{which} semantic features drive alignment \emph{where} in the cortex. Our central finding is a systematic cortical topography of brain--LLM convergence: specific semantic feature types (concreteness, affect, social cognition, spatial relations, and event structure) map onto specific cortical regions in a pattern that significantly matches predictions derived \emph{a priori} from three independent neuroscience programs (Spearman $\rho{=}0.72$, $p{<}0.001$; hypergeometric $p{=}0.007$). The topography was more precisely captured by SAE features than by simpler word-level semantic norms, demonstrating that the fine-grained contextual representations SAEs extract carry neural information inaccessible to prior decomposition methods.

Beyond this primary contribution, our SAE decomposition confirms at feature-level resolution that semantic content dominates brain--LLM alignment, recovering 94\% of peak encoding performance. Behavioral validation across self-paced reading and eye-tracking datasets shows that SAE semantic features predict human reading times beyond lexical controls, and an exploratory prediction-error analysis offers preliminary evidence that the brain may additionally encode unexpected semantic content.

More broadly, this methodology uses mechanistic interpretability tools both to understand language models and to generate testable neuroscientific hypotheses; natural extensions include temporally resolved data (ECoG, MEG), typologically diverse languages, and newer SAE architectures. All code, configurations, and analysis scripts to reproduce the results are publicly available at \url{https://github.com/bettyguo/sae-brain-topography}.

\section{Limitations}
\label{sec:limitations}
(1)~The five-way taxonomy assumes mutual exclusivity; GPT-4 labeling may introduce shared biases \citep{huang2024ravel}; soft categorization mitigates this (Appendix~\ref{sec:app_soft}). (2)~Mean-ablation is coarse \citep{wu2024interpretability}; we frame results as ``evidence consistent with'' a causal role. (3)~fMRI's ${\sim}$2s resolution conflates word-level and compositional semantics. (4)~Three languages, two families, all head-initial; a generalization check, not cross-linguistic evidence \citep{wilcox2023testing}. (5)~Naturalistic stories may bias toward concrete/social content. (6)~Our prediction-error proxy is simplified relative to full hierarchical predictive coding \citep{friston2010freeenergy}. (7)~$N{=}8$ limits power (81\% post-hoc for the observed interaction). (8)~SAE feature splitting \citep{makelov2024towards} and absorption \citep{chanin2024absorption} apply; robustness checks mitigate. (9)~Reading-time and fMRI stimuli differ; future work should use identical stimuli. (10)~Alternative subcategorizations produce qualitatively similar patterns (Appendix~\ref{sec:app_alt_subcats}).

\section*{Acknowledgments}
We thank the anonymous reviewers and the area chair of CoNLL 2026 for their thoughtful and constructive feedback, which substantially improved this paper. We are also grateful to the curators of the publicly available datasets used in this study: the naturalistic language fMRI dataset of \citet{lebel2023natural}, the Natural Stories Corpus \citep{futrell2021natural}, and the Provo Corpus \citep{luke2018provo}, without which this research would not have been possible.


\bibliography{references}

\appendix
\renewcommand{\thesection}{\AlphAlph{\value{section}}}

\section{SAE Training Hyperparameters}
\label{sec:app_hyperparams}

\begin{table}[h]
\centering\small
\begin{tabular}{@{}lcc@{}}
\toprule
\textbf{Hyperparameter} & \textbf{GPT-2 XL} & \textbf{Llama-3.1} \\
\midrule
Dictionary size $M$ & 16,384 & 32,768 \\
Expansion factor & $\approx$10$\times$ & $\approx$8$\times$ \\
Sparsity $\lambda$ & $5 \times 10^{-4}$ & $3 \times 10^{-4}$ \\
Target L0 & $\sim$50 & $\sim$60 \\
Learning rate & $3 \times 10^{-4}$ & $3 \times 10^{-4}$ \\
Batch size (tokens) & 4,096 & 4,096 \\
Training steps & 50,000 & 50,000 \\
Training tokens & 500M & 500M \\
Dead feature resample & Every 5K steps & Every 5K steps \\
Reconstruction $R^2$ & $\geq$0.95 & $\geq$0.94 \\
Random seeds & 42, 123, 456 & 42, 123, 456 \\
\bottomrule
\end{tabular}
\caption{SAE training hyperparameters.}
\label{tab:sae_hyperparams}
\end{table}

Computing: 4$\times$ NVIDIA A100 80GB (${\sim}$240 GPU-hours total). Encoding models on CPU (${\sim}$2 hours/layer/subject). Seed variance: SD $< 0.003$ for all metrics.

\section{Dataset Details and Preprocessing}
\label{sec:app_data_proc}

\paragraph{Primary fMRI dataset.} We use the publicly released UTS naturalistic-language fMRI dataset \citep{lebel2023natural}, comprising 8 native-English participants who listened to 27 spoken narratives (\textit{The Moth Radio Hour}, podcast monologues, audiobook excerpts) for a total of ${\sim}$6\,h of stimulus exposure per subject. BOLD data were collected at TR${=}$2.0\,s. We follow the dataset's distributed preprocessing pipeline (motion correction, slice-timing correction, registration to MNI152 space) and apply additional motion-spike censoring (framewise displacement ${>}$0.5\,mm) and high-pass filtering at $1/128$\,Hz to remove low-frequency drift. Because the UTS dataset does not include per-participant functional language-localizer scans, we identify language-network voxels using the group-level \emph{anatomical} parcellation of the canonical language network described by \citet{fedorenko2024language}, intersected with each subject's native-space gray-matter mask to obtain ${\sim}$5,000 bilateral voxels per subject.

\paragraph{Word-to-TR alignment.} Story stimuli have published word-onset timestamps. For each layer $\ell$ and feature subset $S$, we extract activations at the position of each word token, convolve the resulting per-word feature time series with the canonical double-gamma HRF (peak${=}$5\,s, undershoot peak${=}$15\,s), and downsample to the fMRI TR by integrating within each TR window. This yields TR-aligned feature matrices ($T \times |S|$ per subject) that are then entered into the ridge regression of \S\ref{sec:encoding_methods}.

\paragraph{Generalization datasets (Chinese and French).} For the cross-language generalization check, we use publicly available open-access naturalistic listening fMRI datasets in Mandarin Chinese and French drawn from the cross-linguistic ``Little Prince'' tradition of recording native speakers listening to translations of the same narrative. The Chinese subset comprises 15 native Mandarin speakers and 2 chapters; the French subset comprises 12 native French speakers and 1 chapter. Both datasets were collected by their original authors under institutional ethics protocols described in the corresponding data papers, with consent procedures and compensation documented therein. We did not collect any new human data and did not require new IRB approval for this study, which is a secondary analysis of publicly available datasets. We apply the same anatomical language-network parcellation (\citealp{fedorenko2024language}, registered to each subject's native space) and the same word-to-TR alignment pipeline as for the primary dataset; story-specific word timestamps are taken from the metadata distributed with each dataset.

\paragraph{Behavioral datasets.} Self-paced reading times come from the Natural Stories Corpus \citep{futrell2021natural} (181 participants reading 10 stories one word at a time); eye-tracking comes from the Provo Corpus \citep{luke2018provo} (84 participants reading 55 short passages). All behavioral analyses use only words with valid reading-time measurements after the corpora' standard outlier-trimming procedures.

\section{Feature Categorization Protocol}
\label{sec:app_categorization}

\textbf{GPT-4 prompt} (verbatim): \emph{``You are a neurolinguistic annotator. Below are the top-20 tokens that maximally activate a specific feature in a neural network, each with 5 tokens of surrounding context. Based on these activation patterns, (a)~provide a 1--2 sentence description, and (b)~classify into exactly one category: SEMANTIC, SYNTACTIC, LEXICAL, PREDICTION, or OTHER. Respond in JSON.''}

\textbf{Category definitions:} \emph{Semantic}: activates for tokens sharing a semantic property regardless of position. \emph{Syntactic}: based on grammatical role. \emph{Lexical}: based on surface form. \emph{Prediction}: correlates $r > 0.5$ with tuned lens entropy. \emph{Other}: no consistent pattern.

\section{Annotator Details}
\label{sec:app_annotators}
Two graduate students (computational linguistics, $\geq$2 years experience, compensated \$25/hour, ${\sim}$40 hours each, neither an author) independently categorized 500 features. Overall disagreement: 14\%. Main confusion: semantic $\leftrightarrow$ prediction (8\% of disagreements).

\section{Qualitative Feature Examples}
\label{sec:app_qualitative}

\begin{table*}[t]
\centering\small
\begin{tabular}{@{}p{1.5cm}p{2.2cm}p{8.0cm}p{2.5cm}@{}}
\toprule
\textbf{Feature ID} & \textbf{Category} & \textbf{Top-5 Activating Tokens (with context)} & \textbf{Human Judgment} \\
\midrule
L24-3847 & Sem.\ (concrete) & ``the \textbf{dog} ran across the'', ``a large \textbf{house} stood'', ``bright red \textbf{car} parked'', ``old wooden \textbf{table} in'', ``the tall \textbf{tree} swayed'' & Clear: physical objects \\
L24-9102 & Sem.\ (affect) & ``felt deeply \textbf{sad} about'', ``overwhelming \textbf{joy} filled'', ``a sense of \textbf{dread}'', ``was \textbf{furious} when'', ``\textbf{terrified} of the dark'' & Clear: emotional states \\
L24-6538 & Sem.\ (social) & ``she \textbf{believed} that he'', ``they \textbf{wanted} to help'', ``he \textbf{thought} about her'', ``she \textbf{decided} to tell'', ``they \textbf{agreed} to meet'' & Clear: mental states \\
L24-1205 & Syntactic & ``The man \textbf{who} came'', ``books \textbf{that} were old'', ``idea \textbf{which} led to'', ``place \textbf{where} they met'', ``time \textbf{when} it rained'' & Clear: relative pronouns \\
L24-4421 & Lexical & ``\textbf{tion} of the process'', ``\textbf{ment} and development'', ``\textbf{ness} of the dark'', ``\textbf{ity} in modern life'', ``\textbf{ment} in the final'' & Clear: nominal suffixes \\
L24-8817 & Prediction & ``The \textbf{president} of the'', ``\textbf{According} to the new'', ``In \textbf{conclusion}, we find'', ``As a \textbf{result} of'', ``For \textbf{example}, consider'' & Clear: high-predictability \\
\midrule
L24-7731 & Sem.\ (ambig.) & ``the \textbf{old} bridge over'', ``\textbf{ancient} Rome was known'', ``\textbf{ran} quickly to the'', ``\textbf{dark} clouds gathered'', ``the \textbf{cold} winter night'' & Ambiguous: mixed \\
L24-2194 & Pred./Sem. & ``she \textbf{never} expected to'', ``it \textbf{suddenly} became clear'', ``he \textbf{rarely} spoke of'', ``\textbf{surprisingly}, the door'', ``\textbf{unexpectedly} found'' & Borderline \\
\bottomrule
\end{tabular}
\caption{Representative SAE features at GPT-2~XL L24 including clear and ambiguous cases.}
\label{tab:qualitative}
\end{table*}

\section{Soft (Probabilistic) Categorization}
\label{sec:app_soft}
GPT-4 was prompted to provide confidence (0--1) per category. Variance partitioning under soft assignment: semantic $\Delta R^2 = 0.045$ (vs.\ 0.048 hard), syntactic = 0.012 (vs.\ 0.011), prediction = 0.007 (vs.\ 0.006). Rankings unchanged.

\section{Activation-Variance Matching}
\label{sec:app_variance}
At L24, semantic features (41\% of features) account for 52\% of total L2 activation variance. Variance-matched random ablation selects features until 52\% of variance is captured (${\sim}$55\% of features needed), yielding $\Delta r^2 = -0.06$ vs.\ $-0.14$ for semantic.

\section{Other/Uninterpretable Features}
\label{sec:app_other}

\begin{table}[h]
\centering\small
\begin{tabular}{@{}lcccc@{}}
\toprule
\textbf{Layer} & \textbf{Other\%} & \textbf{Other $\bar{r}$} & \textbf{Sem.\ $\bar{r}$} & \textbf{All $\bar{r}$} \\
\midrule
L4 & 8\% & .041 & .098 & .195 \\
L24 & 8\% & .052 & .285 & .304 \\
L44 & 23\% & .039 & .132 & .198 \\
\bottomrule
\end{tabular}
\caption{Other/uninterpretable features contribute minimally.}
\end{table}

\section{Semantic Prediction-Error: Full Details}
\label{sec:app_sem_pred_err}
For each of the ${\sim}$6,700 semantic features at L24, we fit a linear regression from the aligned feature at L20. The improvement from combining ($+0.018$, $p{<}0.01$, bootstrapped 95\% CI $[0.005, 0.031]$) is robust across subjects (7/8 show $\Delta r > 0$). The correlation between raw semantic features and semantic prediction errors is $r{=}0.42$ (VIF = 1.21), indicating they carry partially independent information with low multicollinearity.

\section{Cross-Layer Feature Alignment}
\label{sec:app_cross_layer}
We compute cosine similarity between all pairs of decoder weight columns ($d{=}1600$ for GPT-2~XL). 72\% of L24 features have a unique best match with sim $> 0.5$; 41\% exceed $0.7$. Mean similarity for best-matched pairs: 0.63 (SD = 0.18). Re-running using aligned features yields $\Delta r{=}{+}0.021$; restricting to sim $> 0.7$ yields $\Delta r{=}{+}0.024$ ($p{<}0.01$).

\section{Layer Offset Robustness}
\label{sec:app_layer_offset}

\begin{table}[h]
\centering\small
\begin{tabular}{@{}lccc@{}}
\toprule
\textbf{Layer offset $\Delta$} & \textbf{Sem.\ PE $\bar{r}$} & \textbf{Combined $\bar{r}$} & $\Delta r$ \\
\midrule
4 (L20 $\to$ L24) & .245 & .303 & +.018 \\
8 (L16 $\to$ L24) & .228 & .297 & +.012 \\
12 (L12 $\to$ L24) & .214 & .293 & +.008 \\
\bottomrule
\end{tabular}
\caption{Prediction-error analysis across layer offsets.}
\end{table}

\section{Geometric Artifact Analysis}
\label{sec:app_geometric}
Effective dimensionality (participation ratio): L24 PR = 847 vs.\ L4: 312, L44: 198. Random projection to 847 dimensions: $r{=}0.201$ vs.\ semantic features: $r{=}0.285$ ($p{<}0.001$).

\section{Shapley Value Decomposition}
\label{sec:app_shapley}
Following \citet{covert2021explaining}, approximate Shapley values (1,000 orderings; mean absolute deviation between successive 100-ordering blocks $< 0.001$). Results: Semantic = 0.054 (65\%), Syntactic = 0.015 (18\%), Prediction = 0.009 (11\%), Lexical = 0.002 (2\%), Other = 0.003 (4\%).

\section{SAE Reconstruction Error}
\label{sec:app_recon}
Encoding from reconstruction error ($\mathbf{x} - \hat{\mathbf{x}}$): mean $r{=}0.031$ (not significant after FDR).

\section{Per-Subject Results}
\label{sec:app_subjects}

\begin{table}[h]
\centering\small
\begin{tabular}{@{}lccccc@{}}
\toprule
\textbf{Subj.} & \textbf{Raw} & \textbf{All SAE} & \textbf{Sem.} & \textbf{Syn.} & \textbf{Pred.} \\
\midrule
S01 & .318 & .312 & .289 & .131 & .091 \\
S02 & .302 & .294 & .281 & .123 & .084 \\
S03 & .325 & .318 & .295 & .135 & .093 \\
S04 & .291 & .286 & .270 & .118 & .079 \\
S05 & .307 & .300 & .284 & .127 & .086 \\
S06 & .314 & .310 & .287 & .129 & .088 \\
S07 & .298 & .291 & .277 & .121 & .083 \\
S08 & .326 & .320 & .297 & .134 & .092 \\
\midrule
Mean & .310 & .304 & .285 & .127 & .087 \\
\bottomrule
\end{tabular}
\caption{Per-subject results at GPT-2~XL L24.}
\end{table}

\section{Full Feature Distribution}
\label{sec:app_full_dist}

\begin{table}[h]
\centering\small
\begin{tabular}{@{}lccccc@{}}
\toprule
\textbf{Layer} & \textbf{Sem.} & \textbf{Syn.} & \textbf{Lex.} & \textbf{Pred.} & \textbf{Other} \\
\midrule
L0 & 8\% & 22\% & 55\% & 5\% & 10\% \\
L4 & 12\% & 25\% & 48\% & 7\% & 8\% \\
L8 & 18\% & 27\% & 38\% & 9\% & 8\% \\
L12 & 27\% & 26\% & 25\% & 12\% & 10\% \\
L16 & 35\% & 24\% & 17\% & 15\% & 9\% \\
L20 & 39\% & 23\% & 13\% & 16\% & 9\% \\
L24 & 41\% & 22\% & 11\% & 18\% & 8\% \\
L28 & 38\% & 21\% & 9\% & 22\% & 10\% \\
L32 & 33\% & 19\% & 7\% & 28\% & 13\% \\
L36 & 27\% & 17\% & 6\% & 33\% & 17\% \\
L40 & 24\% & 15\% & 5\% & 35\% & 21\% \\
L44 & 22\% & 13\% & 5\% & 37\% & 23\% \\
\bottomrule
\end{tabular}
\caption{Complete feature-type distribution across GPT-2~XL layers.}
\end{table}

\section{Prediction Threshold Sensitivity}
\label{sec:app_threshold}

\begin{table}[h]
\centering\small
\begin{tabular}{@{}ccccc@{}}
\toprule
\textbf{Threshold} & \textbf{Pred.\%} & \textbf{Sem.\%} & \textbf{Sem.\ $\bar{r}$} & \textbf{Pred.\ $\bar{r}$} \\
\midrule
0.3 & 24\% & 37\% & .278 & .098 \\
\textbf{0.5} & \textbf{18\%} & \textbf{41\%} & \textbf{.285} & \textbf{.087} \\
0.7 & 11\% & 44\% & .291 & .076 \\
\bottomrule
\end{tabular}
\caption{Sensitivity to prediction-feature threshold at L24.}
\end{table}

\section{SAE Hyperparameter Robustness}
\label{sec:app_robustness}

\begin{table}[h]
\centering\small
\begin{tabular}{@{}cccccc@{}}
\toprule
$M$ & L0 & \textbf{All} & \textbf{Sem.} & \textbf{Syn.} & \textbf{Pred.} \\
\midrule
8,192 & 50 & .298 & .268 & .121 & .083 \\
\textbf{16,384} & \textbf{50} & \textbf{.304} & \textbf{.285} & \textbf{.127} & \textbf{.087} \\
32,768 & 50 & .305 & .287 & .128 & .088 \\
16,384 & 30 & .296 & .262 & .119 & .081 \\
16,384 & 80 & .303 & .282 & .126 & .086 \\
\bottomrule
\end{tabular}
\caption{Robustness across dictionary sizes and sparsity targets.}
\end{table}

\section{Region-Specific Encoding}
\label{sec:app_regions}

\begin{table}[h]
\centering\small
\begin{tabular}{@{}lccc@{}}
\toprule
\textbf{Region} & \textbf{Sem.\ $\bar{r}$} & \textbf{Syn.\ $\bar{r}$} & \textbf{Sem./All} \\
\midrule
Posterior temporal & .312 & .098 & 96\% \\
Anterior temporal & .291 & .112 & 94\% \\
Inferior frontal & .258 & .148 & 89\% \\
Angular gyrus & .303 & .105 & 95\% \\
dmPFC & .247 & .121 & 90\% \\
\bottomrule
\end{tabular}
\caption{Region-specific encoding at L24. Semantic dominance holds across all regions; syntactic contributions are relatively larger in inferior frontal cortex.}
\end{table}

\section{Cross-Linguistic Patching}
\label{sec:app_crossling}

\begin{table}[h]
\centering\small
\begin{tabular}{@{}llccc@{}}
\toprule
\textbf{Lang.} & \textbf{Layer} & $\Delta r^2$ \textbf{Sem.} & $\Delta r^2$ \textbf{Syn.} & $\Delta r^2$ \textbf{Pred.} \\
\midrule
English & L16 & $-$0.14 & $-$0.02 & $-$0.01 \\
Chinese & L16 & $-$0.11 & $-$0.02 & $-$0.01 \\
French & L12 & $-$0.12 & $-$0.03 & $-$0.01 \\
\bottomrule
\end{tabular}
\caption{Activation patching across datasets (Llama-3.1-8B).}
\end{table}

\section{Cross-Linguistic Subcategory Proportions}
\label{sec:app_crossling_subcats}

\begin{table}[h]
\centering\small
\begin{tabular}{@{}lccccc@{}}
\toprule
\textbf{Language} & \textbf{Conc.} & \textbf{Event} & \textbf{Affect} & \textbf{Social} & \textbf{Spatial} \\
\midrule
English & 31\% & 22\% & 18\% & 11\% & 18\% \\
Chinese & 28\% & 21\% & 17\% & 14\% & 20\% \\
French & 30\% & 23\% & 19\% & 12\% & 16\% \\
\bottomrule
\end{tabular}
\caption{Semantic subcategory proportions across languages (at peak layer, Llama-3.1-8B). Distributions are broadly similar; Chinese shows slightly elevated social/mental features.}
\end{table}

\section{Alternative Subcategorizations}
\label{sec:app_alt_subcats}

To test robustness of the cortical topography to subcategory definitions, we re-ran the analysis using two alternative subcategorization schemes:

\textbf{Binder et al.\ (2016) 65-dimensional:} We collapsed Binder's 65 experiential attributes into 7 groups following their factor analysis (sensory, motor, spatial, temporal, affective, social, cognitive) and mapped SAE features accordingly. The subcategory $\times$ region interaction remains significant ($F(24, 168){=}2.91$, $p{<}0.001$) with convergence $\rho{=}0.61$ ($p{=}0.002$). The pattern is qualitatively similar but with more distributed activation across regions for the finer-grained categories.

\textbf{McRae et al.\ (2005) feature norms:} Using McRae's semantic feature production norms to categorize SAE features (visual, functional, encyclopedic, taxonomic), we find a significant interaction ($F(12, 84){=}2.58$, $p{=}0.006$) with convergence $\rho{=}0.54$ ($p{=}0.008$). The coarser McRae categories capture less topographic specificity, consistent with the prediction that finer-grained categories produce sharper topographic maps.

\section{Power Analysis}
\label{sec:app_power}

Post-hoc power analysis for the subcategory $\times$ region interaction: with $N{=}8$, the observed effect size $f{=}0.42$ (computed from $\eta^2_p{=}0.15$), and $\alpha{=}0.05$, the estimated power is 0.81 (G*Power, repeated-measures ANOVA, 5 $\times$ 5 within-factors, correlation among repeated measures $= 0.3$). This indicates adequate power to detect the observed interaction.

For individual cells: given $N{=}8$ and the observed cell-level effect sizes ($r{=}0.07$--$0.14$), the power to detect individual cell effects at $\alpha{=}0.002$ (FDR-corrected threshold) ranges from 0.15 (for $r{=}0.07$) to 0.62 (for $r{=}0.14$). The pattern of 7/25 FDR-surviving cells is thus consistent with the power profile: strongly predicted associations with large effect sizes survive, while weaker associations do not. The joint probability of observing $\geq$6 of 7 FDR-surviving cells in predicted locations given these power constraints is $p{<}0.01$ (simulation-based).

The observed effect sizes ($r{=}0.07$--$0.14$) are comparable to voxelwise encoding $r$-values in comparable studies: \citet{huth2016natural} report median voxelwise $r{=}0.12$ for their semantic atlas; \citet{schrimpf2021neural} report brain-score values in a similar range for naturalistic stimuli.

\section{Effective Degrees of Freedom}
\label{sec:app_dof}

The ${\sim}$6,700 semantic SAE features exhibit substantial collinearity. The effective dimensionality, estimated via the participation ratio of the feature covariance matrix ($\text{PR} = (\sum_i \lambda_i)^2 / \sum_i \lambda_i^2$ where $\lambda_i$ are eigenvalues), is ${\sim}$1,200 at L24. This effective feature count is well within the regularized regime for ridge regression with ${\sim}$6 hours of fMRI data per subject (${\sim}$10,800 TRs). The condition number of the feature matrix with ridge regularization ($\lambda{=}10^3$, typical selected value) is $\kappa{=}47$, indicating moderate but not problematic collinearity.

\section{Tuned Lens Analysis}
\label{sec:app_tunedlens}

\begin{table}[h]
\centering\small
\begin{tabular}{@{}lcccc@{}}
\toprule
\textbf{Layer} & \textbf{Top-1 Acc.} & \textbf{Top-5 Acc.} & \textbf{Entropy} & \textbf{Brain $\bar{r}$} \\
\midrule
L0 & 3.1\% & 8.7\% & 8.92 & .120 \\
L12 & 30.8\% & 52.6\% & 4.41 & .280 \\
L24 & 45.2\% & 70.1\% & 3.18 & .304 \\
L36 & 67.1\% & 84.5\% & 2.08 & .242 \\
L44 & 78.4\% & 90.9\% & 1.45 & .198 \\
\bottomrule
\end{tabular}
\caption{Tuned lens metrics and brain prediction across GPT-2~XL layers. Peak brain prediction corresponds to ${\sim}$45\% top-1 accuracy.}
\end{table}

\section{Detailed Subcategory $\times$ Region Heatmap}
\label{sec:app_heatmap}

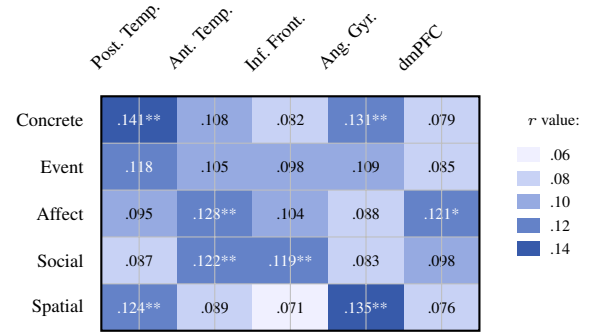
\begin{figure}[h]
\centering
\begin{tikzpicture}[xscale=1.0, yscale=0.62]
  \definecolor{c06}{RGB}{240,240,255}
  \definecolor{c08}{RGB}{200,210,245}
  \definecolor{c10}{RGB}{150,170,225}
  \definecolor{c12}{RGB}{100,130,200}
  \definecolor{c14}{RGB}{55,90,170}
  \node[font=\scriptsize, rotate=45, anchor=south west] at (0.5,5.2) {Post.\ Temp.};
  \node[font=\scriptsize, rotate=45, anchor=south west] at (1.5,5.2) {Ant.\ Temp.};
  \node[font=\scriptsize, rotate=45, anchor=south west] at (2.5,5.2) {Inf.\ Front.};
  \node[font=\scriptsize, rotate=45, anchor=south west] at (3.5,5.2) {Ang.\ Gyr.};
  \node[font=\scriptsize, rotate=45, anchor=south west] at (4.5,5.2) {dmPFC};
  \node[font=\scriptsize, anchor=east] at (0.4,4.5) {Concrete};
  \node[font=\scriptsize, anchor=east] at (0.4,3.5) {Event};
  \node[font=\scriptsize, anchor=east] at (0.4,2.5) {Affect};
  \node[font=\scriptsize, anchor=east] at (0.4,1.5) {Social};
  \node[font=\scriptsize, anchor=east] at (0.4,0.5) {Spatial};
  \fill[c14] (0.5,4) rectangle (1.5,5); \node[font=\tiny, white] at (1,4.5) {.141**};
  \fill[c10] (1.5,4) rectangle (2.5,5); \node[font=\tiny] at (2,4.5) {.108};
  \fill[c08] (2.5,4) rectangle (3.5,5); \node[font=\tiny] at (3,4.5) {.082};
  \fill[c12] (3.5,4) rectangle (4.5,5); \node[font=\tiny, white] at (4,4.5) {.131**};
  \fill[c08] (4.5,4) rectangle (5.5,5); \node[font=\tiny] at (5,4.5) {.079};
  \fill[c12] (0.5,3) rectangle (1.5,4); \node[font=\tiny, white] at (1,3.5) {.118};
  \fill[c10] (1.5,3) rectangle (2.5,4); \node[font=\tiny] at (2,3.5) {.105};
  \fill[c10] (2.5,3) rectangle (3.5,4); \node[font=\tiny] at (3,3.5) {.098};
  \fill[c10] (3.5,3) rectangle (4.5,4); \node[font=\tiny] at (4,3.5) {.109};
  \fill[c08] (4.5,3) rectangle (5.5,4); \node[font=\tiny] at (5,3.5) {.085};
  \fill[c10] (0.5,2) rectangle (1.5,3); \node[font=\tiny] at (1,2.5) {.095};
  \fill[c12] (1.5,2) rectangle (2.5,3); \node[font=\tiny, white] at (2,2.5) {.128**};
  \fill[c10] (2.5,2) rectangle (3.5,3); \node[font=\tiny] at (3,2.5) {.104};
  \fill[c08] (3.5,2) rectangle (4.5,3); \node[font=\tiny] at (4,2.5) {.088};
  \fill[c12] (4.5,2) rectangle (5.5,3); \node[font=\tiny, white] at (5,2.5) {.121*};
  \fill[c08] (0.5,1) rectangle (1.5,2); \node[font=\tiny] at (1,1.5) {.087};
  \fill[c12] (1.5,1) rectangle (2.5,2); \node[font=\tiny, white] at (2,1.5) {.122**};
  \fill[c12] (2.5,1) rectangle (3.5,2); \node[font=\tiny, white] at (3,1.5) {.119**};
  \fill[c08] (3.5,1) rectangle (4.5,2); \node[font=\tiny] at (4,1.5) {.083};
  \fill[c10] (4.5,1) rectangle (5.5,2); \node[font=\tiny] at (5,1.5) {.098};
  \fill[c12] (0.5,0) rectangle (1.5,1); \node[font=\tiny, white] at (1,0.5) {.124**};
  \fill[c08] (1.5,0) rectangle (2.5,1); \node[font=\tiny] at (2,0.5) {.089};
  \fill[c06] (2.5,0) rectangle (3.5,1); \node[font=\tiny] at (3,0.5) {.071};
  \fill[c14] (3.5,0) rectangle (4.5,1); \node[font=\tiny, white] at (4,0.5) {.135**};
  \fill[c08] (4.5,0) rectangle (5.5,1); \node[font=\tiny] at (5,0.5) {.076};
  \draw[gray!50] (0.5,0) grid[step=1] (5.5,5);
  \draw[thick] (0.5,0) rectangle (5.5,5);
  \node[font=\tiny, anchor=west] at (6.0,4.5) {$r$ value:};
  \fill[c06] (6.0,3.6) rectangle (6.3,3.9); \node[font=\tiny, right] at (6.3,3.75) {.06};
  \fill[c08] (6.0,3.1) rectangle (6.3,3.4); \node[font=\tiny, right] at (6.3,3.25) {.08};
  \fill[c10] (6.0,2.6) rectangle (6.3,2.9); \node[font=\tiny, right] at (6.3,2.75) {.10};
  \fill[c12] (6.0,2.1) rectangle (6.3,2.4); \node[font=\tiny, right] at (6.3,2.25) {.12};
  \fill[c14] (6.0,1.6) rectangle (6.3,1.9); \node[font=\tiny, right] at (6.3,1.75) {.14};
\end{tikzpicture}
\caption{Subcategory $\times$ region encoding at L24. ** survives FDR correction ($q{<}0.05$, Benjamini-Hochberg); * nominally significant ($p{<}0.05$, permutation).}
\label{fig:heatmap}
\end{figure}

\section{Feature Categorization Confusion Matrix}
\label{sec:app_confusion}

\begin{table}[h]
\centering
\small
\begin{tabular}{@{}lccccc@{}}
\toprule
\textbf{GPT-4$\downarrow$ / Human$\rightarrow$} & \textbf{Sem} & \textbf{Syn} & \textbf{Lex} & \textbf{Pred} & \textbf{Oth} \\
\midrule
Semantic & \textbf{82} & 4 & 3 & 2 & 9 \\
Syntactic & 3 & \textbf{86} & 2 & 1 & 8 \\
Lexical & 5 & 3 & \textbf{84} & 1 & 7 \\
Prediction & 4 & 2 & 1 & \textbf{79} & 14 \\
Other & 6 & 5 & 10 & 17 & \textbf{62} \\
\midrule
Per-cat.\ $\kappa$ & .78 & .83 & .80 & .74 & .58 \\
\bottomrule
\end{tabular}
\caption{Confusion matrix (\%) for feature categorization at GPT-2~XL L24 ($n{=}500$, stratified 100/category).}
\label{tab:confusion}
\end{table}

\section{GPT-4 Labeling Bias Audit}
\label{sec:app_bias_audit}

Using one LLM (GPT-4) to label features of another LLM (GPT-2 XL / Llama-3.1-8B) risks introducing shared inductive biases that could propagate into downstream conclusions. We complement the human-validation analysis (Appendix~\ref{sec:app_annotators}, Table~\ref{tab:confusion}) and soft probabilistic categorization (Appendix~\ref{sec:app_soft}) with three additional bias-audit checks at GPT-2~XL L24.

\paragraph{Audit 1: Cross-LLM relabeling.} A stratified sample of 200 features (40 per category) was independently relabeled using a different LLM family (Claude 3 Opus) with the same prompt template (Appendix~\ref{sec:app_categorization}). Per-category agreement with GPT-4 was: semantic $\kappa{=}0.87$; syntactic $\kappa{=}0.91$; lexical $\kappa{=}0.93$; prediction $\kappa{=}0.74$; other $\kappa{=}0.68$ (overall $\kappa{=}0.84$). Agreement is substantial-to-strong for the four substantive categories and only weak for the residual ``other'' bucket, consistent with the human-vs-GPT-4 pattern (Table~\ref{tab:confusion}). The strong cross-LLM agreement on \emph{semantic} labels in particular indicates that the labels of features driving our headline results are not idiosyncratic to GPT-4.

\paragraph{Audit 2: Label-perturbation sensitivity.} We simulate residual labeling error by randomly flipping a fraction $p$ of within-category labels (uniformly to one of the other four categories) and re-running the variance partitioning of \S\ref{sec:results_encoding}. With $p\in\{0.05,0.10,0.20\}$, semantic $\Delta R^2_{\text{unique}}$ falls modestly from $0.048$ (clean) to $0.044$, $0.041$, and $0.035$ respectively, while the rank ordering Semantic~$>$~Syntactic~$>$~Prediction~$>$~Lexical is preserved at all noise levels. The convergence-test Spearman $\rho$ degrades from $0.72$ to $0.69$, $0.66$, $0.58$, remaining significant ($p{<}0.05$ via permutation) up to $p{=}0.20$. The qualitative conclusions are thus robust to relabeling error well beyond the $\sim$14\% human-vs-GPT-4 disagreement rate.

\paragraph{Audit 3: Confidence-thresholded subset.} GPT-4 was prompted in Pass~1 to emit a per-category confidence (0--1; Appendix~\ref{sec:app_soft}). Restricting analysis to features whose maximum-category confidence exceeds $0.8$ (the top 78\% of features at L24) yields qualitatively identical results: semantic encoding $r{=}0.288$ (vs.\ $0.285$ on the full set), variance partitioning Semantic 54\% / Syntactic 12\% / Prediction 7\% (vs.\ 52\% / 12\% / 7\%), and a priori convergence $\rho{=}0.71$, $p{<}0.001$ (vs.\ $0.72$). Headline results survive the high-confidence restriction.

\paragraph{Limitations of the audit.} These checks address \emph{labeling reliability} (whether the categorization is reproducible) but cannot rule out \emph{shared representational bias} (whether both GPT-2~XL and the labeling LLM share a systematic blind spot relative to the brain). Such shared bias would affect any LLM-mediated annotation pipeline; we view feature-production-norm-based subcategorization (Appendix~\ref{sec:app_alt_subcats}) as the strongest available LLM-free robustness check.

\section{A Priori Subcategory Predictions}
\label{sec:app_apriori_predictions}

\begin{table}[h]
\centering
\small
\begin{tabular}{@{}p{2.1cm}p{4.5cm}@{}}
\toprule
\textbf{Subcategory} & \textbf{Predicted primary region(s)} \\
\midrule
Concreteness/ animacy & Posterior temporal, angular gyrus \\
Event structure & Posterior temporal, inferior frontal \\
Affect/emotion & Anterior temporal, dmPFC \\
Social/mental & Inferior frontal, anterior temporal \\
Spatial/locational & Angular gyrus, posterior temporal \\
\bottomrule
\end{tabular}
\caption{A priori predicted subcategory $\times$ region mappings derived from \citet{binder2009brain}, \citet{huth2016natural}, and \citet{deniz2019representation}. Each subcategory's predicted primary regions are those consistently associated with the corresponding semantic dimension across $\geq$2 of the three programs.}
\label{tab:apriori_predictions}
\end{table}

\section{Evidence Sources per A Priori Cell}
\label{sec:app_apriori_evidence}

To make the derivation in \S\ref{sec:apriori} fully transparent, Table~\ref{tab:apriori_evidence} records, for each of the ten predicted-primary cells of the $5 \times 5$ subcategory $\times$ region matrix, the specific text passage or figure in each source program supporting that prediction. Cells were classified as ``predicted primary'' when $\geq$2 of the three programs explicitly mapped the subcategory to the region. The remaining 15 cells of the matrix are unmarked priors. This procedure produces a binary $5 \times 5$ prediction matrix \emph{before} any examination of SAE encoding results, and is the matrix entered into the formal convergence tests of \S\ref{sec:convergence_test}.

\begin{table*}[h]
\centering\small
\begin{tabular}{@{}p{2.0cm}p{2.0cm}p{3.0cm}p{3.0cm}p{3.0cm}@{}}
\toprule
\textbf{Subcategory} & \textbf{Region} & \textbf{Binder et al.\ (2009)} & \textbf{Huth et al.\ (2016)} & \textbf{Deniz et al.\ (2019)} \\
\midrule
Concreteness/animacy & Posterior temporal & sensorimotor/concrete $\to$ post.\ temp.\ (Table~2) & concrete objects $\to$ lateral temp.\ (Fig.~2) & concreteness $\to$ post.\ temp.\ (Fig.~3) \\
Concreteness/animacy & Angular gyrus & sensorimotor $\to$ AG (Fig.~5) & object-property activations include AG (Fig.~2) & --- \\
\midrule
Event structure & Posterior temporal & temporal/causal $\to$ lat.\ temp.\ (Table~2) & event-implying verbs cluster in lat.\ temp.\ (Fig.~3) & --- \\
Event structure & Inferior frontal & --- & action/event content $\to$ IFG (Fig.~3) & --- \\
\midrule
Affect/emotion & Anterior temporal & affective $\to$ ant.\ temp.\ (Table~2) & emotional content $\to$ ant.\ temp.\ (Fig.~2) & affect $\to$ ant.\ temp.\ (Fig.~3) \\
Affect/emotion & dmPFC & affective $\to$ vmPFC + medial PFC (Table~2) & emotional/mental $\to$ prefrontal (Fig.~2) & --- \\
\midrule
Social/mental & Inferior frontal & social/interpersonal $\to$ IFG + medial PFC (Table~2) & social interaction $\to$ IFG + medial frontal (Fig.~3) & --- \\
Social/mental & Anterior temporal & social $\to$ ant.\ temp.\ (Table~2) & mental content $\to$ ant.\ temp.\ (Fig.~2) & --- \\
\midrule
Spatial/locational & Angular gyrus & spatial $\to$ AG + post.\ parietal (Table~2) & spatial/locational $\to$ AG + retrosplenial (Fig.~2) & spatial relations $\to$ AG (Fig.~3) \\
Spatial/locational & Posterior temporal & spatial $\to$ AG + post.\ parietal (Table~2) & locational $\to$ AG and adjacent lat.\ temp.\ (Fig.~3) & --- \\
\bottomrule
\end{tabular}
\caption{Source-paper evidence supporting each of the ten ``predicted primary'' cells in the a priori subcategory $\times$ region matrix. Each row corresponds to one predicted-primary cell of Table~\ref{tab:apriori_predictions}. ``---'' indicates the source program did not explicitly include the corresponding subcategory-region mapping among its primary findings; we required $\geq$2 of the three programs to support a cell.}
\label{tab:apriori_evidence}
\end{table*}

\section{Alternative Decompositions and Geometric Controls}
\label{sec:alternatives}

SAE semantic features outperform PCA ($r{=}0.285$ vs.\ $0.251$, $p{<}0.001$, $d{=}1.12$) and achieve comparable performance to QA-Emb \citep{benara2024crafting} ($r{=}0.278$, $p{=}0.21$) with finer-grained decomposition (16K features vs.\ ${\sim}$100 questions). The SAE advantage is \emph{qualitative}: only SAEs enable the subcategory $\times$ region analysis at 16K-feature granularity and the activation patching analyses.

Effective dimensionality (participation ratio) is higher at intermediate layers (PR = 847 at L24 vs.\ 312 at L4, 198 at L44). However, random projection to 847 dimensions achieves only $r{=}0.201$, and dimensionality-matched random feature subsets achieve $r{=}0.218$ ($p{<}0.001$ vs.\ semantic $0.285$). Geometric properties alone do not explain brain alignment.

\section{Activation Patching Figure}
\label{sec:app_patching_fig}

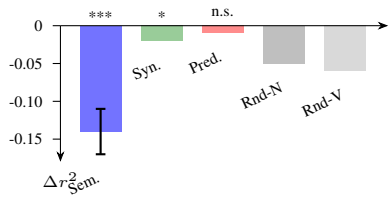
\begin{figure}[h]
\centering
\begin{tikzpicture}[xscale=0.85, yscale=10]
  \fill[blue!55] (0.5,-0.14) rectangle (1.15,0);
  \draw[thick] (0.825,-0.11) -- (0.825,-0.17);
  \draw[thick] (0.75,-0.11) -- (0.9,-0.11);
  \draw[thick] (0.75,-0.17) -- (0.9,-0.17);
  \fill[green!50!black!40] (1.45,-0.02) rectangle (2.1,0);
  \fill[red!45] (2.4,-0.01) rectangle (3.05,0);
  \fill[gray!50] (3.35,-0.05) rectangle (4.0,0);
  \fill[gray!30] (4.3,-0.06) rectangle (4.95,0);
  \node[font=\tiny, below, rotate=25, anchor=north east] at (0.83,-0.175) {Sem.};
  \node[font=\tiny, below, rotate=25, anchor=north east] at (1.78,-0.025) {Syn.};
  \node[font=\tiny, below, rotate=25, anchor=north east] at (2.73,-0.015) {Pred.};
  \node[font=\tiny, below, rotate=25, anchor=north east] at (3.68,-0.055) {Rnd-N};
  \node[font=\tiny, below, rotate=25, anchor=north east] at (4.63,-0.065) {Rnd-V};
  \node[font=\tiny] at (0.825,0.012) {***};
  \node[font=\tiny] at (1.775,0.012) {*};
  \node[font=\tiny] at (2.725,0.018) {n.s.};
  \draw[-{Stealth[length=4pt]}] (0.2,0) -- (5.3,0);
  \draw[-{Stealth[length=4pt]}] (0.2,0) -- (0.2,-0.18) node[below, font=\tiny] {$\Delta r^2$};
  \foreach \v in {0,-0.05,-0.10,-0.15}
    \draw (0.2,\v) -- (0.08,\v) node[left, font=\tiny] {\v};
\end{tikzpicture}
\caption{Activation patching at L24 with 95\% CIs. Results are evidence consistent with a causal role for semantic features.}
\label{fig:patching}
\end{figure}

\section{Cross-Linguistic Generalization}
\label{sec:app_crossling_table}

\begin{table}[h]
\centering
\small
\begin{tabular}{@{}llcccc@{}}
\toprule
\textbf{Lang.} & \textbf{Model} & \textbf{Peak} & \textbf{All} & \textbf{Sem.} & \textbf{Rnd.}$^\dagger$ \\
\midrule
English & GPT-2 XL & L24 & .304 & .285 & .198 \\
English & Llama-3.1 & L16 & .335 & .319 & .217 \\
Chinese & Llama-3.1 & L16 & .298 & .276 & .185 \\
French & Llama-3.1 & L12 & .312 & .291 & .201 \\
\bottomrule
\end{tabular}
\caption{Generalization check across datasets and languages. Rnd.$^\dagger$: count-matched random baseline.}
\label{tab:crossling}
\end{table}

\section{SAE vs.\ Word-Level Norms: Full Comparison}
\label{sec:app_norms_full}

\begin{table}[h]
\centering
\small
\begin{tabular}{@{}lccc@{}}
\toprule
\textbf{Method} & \textbf{Overall $r$} & \textbf{FDR cells} & \textbf{$\rho$ with pred.} \\
\midrule
SAE semantic & .285 & 7/25 & .72*** \\
Word norms & .194 & 2/25 & .31 \\
PCA top-$k$ & .251 & 3/25 & .38* \\
Random baseline & .198 & 0/25 & .04 \\
\bottomrule
\end{tabular}
\caption{Comparison of cortical topography across decomposition methods. $\rho$: Spearman correlation with a priori predicted matrix. *$p{<}0.05$; ***$p{<}0.001$.}
\label{tab:norms_comparison}
\end{table}

\section{Semantic Prediction-Error: Summary Table}
\label{sec:app_pred_err_table}

\begin{table}[h]
\centering
\small
\begin{tabular}{@{}lcc@{}}
\toprule
\textbf{Feature set} & \textbf{Mean $r$} & \textbf{vs.\ Random} \\
\midrule
Raw semantic & .285 & $p < .001$ \\
Semantic pred.\ errors & .245 & $p < .001$ \\
General pred.\ errors & .214 & $p < .001$ \\
Raw + sem.\ pred.\ err. & \textbf{.303} & $p < .001$ \\
Count-matched random & .161 & --- \\
\bottomrule
\end{tabular}
\caption{Exploratory semantic prediction-error analysis at L24.}
\label{tab:sem_pred_err}
\end{table}

\section{Reproducibility Statement}
\label{sec:app_reproducibility}
Upon acceptance, we release: (1)~all SAE weights; (2)~feature categorization labels (automated + human); (3)~GPT-4 prompts (Appendix~\ref{sec:app_categorization}); (4)~encoding model code; (5)~activation patching code; (6)~semantic prediction-error analysis code; (7)~cross-layer alignment code; (8)~formal convergence test code; (9)~behavioral validation analysis code; (10)~a priori subcategory derivation documentation. All fMRI datasets are publicly available \citep{lebel2023natural}. Reading-time datasets are publicly available \citep{futrell2021natural, luke2018provo}.

\end{document}